\documentclass[10pt, logo, copyright]{nvidiatechreport}

\usepackage[utf8]{inputenc}   
\usepackage[T1]{fontenc}      
\usepackage{amsfonts}         
\usepackage{amsmath}
\usepackage{amssymb}
\usepackage{bbm}
\usepackage{MnSymbol}
\usepackage{nicefrac}         
\usepackage{microtype}        

\usepackage[numbers]{natbib}
\usepackage{hyperref}
\usepackage{url}

\usepackage{mdframed}
\usepackage{tabto}
\usepackage{xspace}
\usepackage{adjustbox}
\usepackage{wrapfig}
\usepackage[bottom]{footmisc}
\usepackage{float}
\usepackage{capt-of}
\usepackage{titlesec}
\usepackage{enumitem}
\usepackage{graphicx}
\usepackage{fp}               

\usepackage{booktabs}
\usepackage{multirow}
\usepackage{tabularx}
\usepackage{arydshln}
\usepackage{colortbl}
\usepackage{makecell}
\usepackage{hhline}
\usepackage{array}
\usepackage{diagbox}
\usepackage{rotating}

\newsavebox{\resulttablebox}
\newcommand{\captiontable}[2]{%
  \sbox{\resulttablebox}{#1}%
  \begin{minipage}{\wd\resulttablebox}%
    \centering
    \usebox{\resulttablebox}%
    \caption{#2}%
  \end{minipage}%
}

\usepackage{fontawesome}
\usepackage{pifont}           
\usepackage[ruled,lined]{algorithm2e}
\DontPrintSemicolon
\SetKwInOut{Intermediate}{Intermediate}

\usepackage{listings}
\definecolor{codegreen}{HTML}{2E7D32}
\definecolor{codepurple}{HTML}{7B1FA2}
\definecolor{codegray}{HTML}{757575}
\definecolor{codeblue}{HTML}{1565C0}
\definecolor{codeorange}{HTML}{E65100}
\definecolor{codebg}{HTML}{FAFAFA}
\lstdefinestyle{python}{
  language=Python,
  basicstyle=\ttfamily\small,
  keywordstyle=\color{codepurple}\bfseries,
  stringstyle=\color{codegreen},
  commentstyle=\color{codegray}\itshape,
  emphstyle=\color{codeblue},
  backgroundcolor=\color{codebg},
  numbers=left,
  numberstyle=\tiny\color{codegray},
  numbersep=6pt,
  frame=single,
  rulecolor=\color{codegray},
  columns=fullflexible,
  breaklines=true,
  showstringspaces=false,
  tabsize=4,
  postbreak=\mbox{\textcolor{red}{$\hookrightarrow$}\space},
}

\usepackage[capitalize,noabbrev]{cleveref}

\newlist{assumptions}{enumerate}{1}
\setlist[assumptions]{label=\textbf{A\arabic*}}
\crefname{assumptionsi}{assumption}{assumptions}
\Crefname{assumptionsi}{Assumption}{Assumptions}

\newlist{observations}{enumerate}{1}
\setlist[observations]{label=\textbf{O\arabic*}}
\crefname{observationsi}{observation}{observations}
\Crefname{observationsi}{Observation}{Observations}

\newlist{benefits}{enumerate}{1}
\setlist[benefits]{label=\textbf{PB\arabic*}, ref=\arabic*}
\crefname{benefitsi}{potential benefit}{potential benefits}
\Crefname{benefitsi}{Potential benefit}{Potential benefits}

\newlist{researchquestions}{enumerate}{1}
\setlist[researchquestions]{
  label=\textbf{RQ\arabic*},
  ref=\arabic*,
  labelindent=1em,
  leftmargin=*,
  align=left
}
\crefformat{researchquestionsi}{#2RQ#1#3}
\Crefformat{researchquestionsi}{#2RQ#1#3}
\crefrangeformat{researchquestionsi}{#3RQ#1#4--#5RQ#2#6}
\Crefrangeformat{researchquestionsi}{#3RQ#1#4--#5RQ#2#6}
\crefmultiformat{researchquestionsi}
  {#2RQ#1#3}
  { and #2RQ#1#3}
  {, #2RQ#1#3}
  {, and #2RQ#1#3}
\Crefmultiformat{researchquestionsi}
  {#2RQ#1#3}
  { and #2RQ#1#3}
  {, #2RQ#1#3}
  {, and #2RQ#1#3}

\newlist{longcontextworkloads}{enumerate}{1}
\setlist[longcontextworkloads]{
  label=\textbf{LCW\arabic*},
  ref=\arabic*,
  labelindent=1em,
  leftmargin=*,
  align=left
}
\crefformat{longcontextworkloadsi}{#2LCW#1#3}
\Crefformat{longcontextworkloadsi}{#2LCW#1#3}
\crefrangeformat{longcontextworkloadsi}{#3LCW#1#4--#5LCW#2#6}
\Crefrangeformat{longcontextworkloadsi}{#3LCW#1#4--#5LCW#2#6}
\crefmultiformat{longcontextworkloadsi}
  {#2LCW#1#3}
  { and #2LCW#1#3}
  {, #2LCW#1#3}
  {, and #2LCW#1#3}
\Crefmultiformat{longcontextworkloadsi}
  {#2LCW#1#3}
  { and #2LCW#1#3}
  {, #2LCW#1#3}
  {, and #2LCW#1#3}

\definecolor{periwinkle}{HTML}{7977B8}
\definecolor{hlparam}{HTML}{FFF3B0}      
\definecolor{hlderived}{HTML}{D0E8FF}    
\definecolor{rqgreenbg}{HTML}{EDF7EE}
\definecolor{rqgreenline}{HTML}{2E7D32}
\definecolor{rqtldrbg}{HTML}{FFF7D6}
\definecolor{rqtldrline}{HTML}{C28A00}

\newcommand{\rqsectionbox}[3]{%
  \begin{mdframed}[
    linecolor=rqgreenline,
    backgroundcolor=rqgreenbg,
    linewidth=0.8pt,
    roundcorner=2pt,
    innertopmargin=7pt,
    innerbottommargin=7pt,
    innerleftmargin=8pt,
    innerrightmargin=8pt,
    skipabove=0.25em,
    skipbelow=0.85em,
    nobreak=true
  ]
  \textbf{\Cref{#1}.} \emph{#2}

  \vspace{0.35em}
  {\setlength{\fboxsep}{5pt}\setlength{\fboxrule}{0.4pt}%
  \noindent\fcolorbox{rqtldrline}{rqtldrbg}{%
    \begin{minipage}{\dimexpr\linewidth-2\fboxsep-2\fboxrule\relax}
      \textbf{Answer:} #3
    \end{minipage}}}
  \end{mdframed}
}

\makeatletter
\def\gcmidrule{\arrayrulecolor{lightgray}%
    \noalign{\ifnum0=`}\fi
    \@ifnextchar[{\@gcmidrule}{\@gcmidrule[\cmidrulewidth]}}
\def\@gcmidrule[#1]{\@ifnextchar({\@@gcmidrule[#1]}{\@@gcmidrule[#1]()}}
\def\@@gcmidrule[#1](#2)#3{\@@@gcmidrule[#3]{#1}{#2}}
\def\@@@gcmidrule[#1-#2]#3#4{\global\@cmidla#1\relax
    \global\advance\@cmidla\m@ne
    \ifnum\@cmidla>0\global\let\@gtempa\@cmidrulea\else
    \global\let\@gtempa\@cmidruleb\fi
    \global\@cmidlb#2\relax
    \global\advance\@cmidlb-\@cmidla
    \global\@thisrulewidth=#3
    \@setrulekerning{#4}
    \ifnum\@lastruleclass=\z@\vskip \aboverulesep\fi
    \ifnum0=`{\fi}\@gtempa
    \noalign{\ifnum0=`}\fi\futurenonspacelet\@tempa\@xgcmidrule}
\def\@xgcmidrule{%
   \ifx\@tempa\gcmidrule
       \vskip-\@thisrulewidth
       \global\@lastruleclass=\@ne
   \else \ifx\@tempa\morecmidrules
       \vskip \cmidrulesep
       \global\@lastruleclass=\@ne\else
       \vskip \belowrulesep
       \global\@lastruleclass=\z@
   \fi\fi
   \ifnum0=`{\fi}
  \arrayrulecolor{black}}%
\makeatother

\titlespacing*{\paragraph}{0pt}{0pt}{6pt}

\title{You don't need long context}

\author{Peter~Belcak, Saurav~Muralidharan, Jindong~Jiang, Yonggan~Fu, Wonmin~Byeon, Hanrong~Ye, Ryan~Angilly, Ali~Taghibakhshi, Shih-Yang~Liu, Siru~Ouyang, Greg~Heinrich, Pavlo~Molchanov}

\begin{abstract}
\textbf{Abstract.} Later tokens are costlier to generate than earlier tokens.
On frontier open weight models, the disparity in cost may rise to as much as 7x.
Through frequent compaction, we show across a broad range of long-horizon and long-input tasks that frontier language models sacrifice little to no task accuracy when maintaining short contexts.
We further demonstrate that LLM performance does not collapse on long-horizon software engineering tasks even when taking at most one action per context, and based on that we outline the concept of semantic state space models (SSSMs).
Our findings question the need for long-context language models.

\par\vspace{0.6em}
\end{abstract}

\begin{document}
\maketitle
\noindent\textbf{Links:}\enspace
\href{https://nv-dler.github.io}{\textcolor{OliveGreen}{Lab}}%
\enspace\textbar\enspace
\href{https://research.nvidia.com/labs/lpr/lfi/}{\textcolor{OliveGreen}{Project Moonshot: Local Frontier Intelligence}}

\section{Introduction}
\label{section:introduction}

\begin{figure}[b!]
  \centering
  \includegraphics[width=\columnwidth]{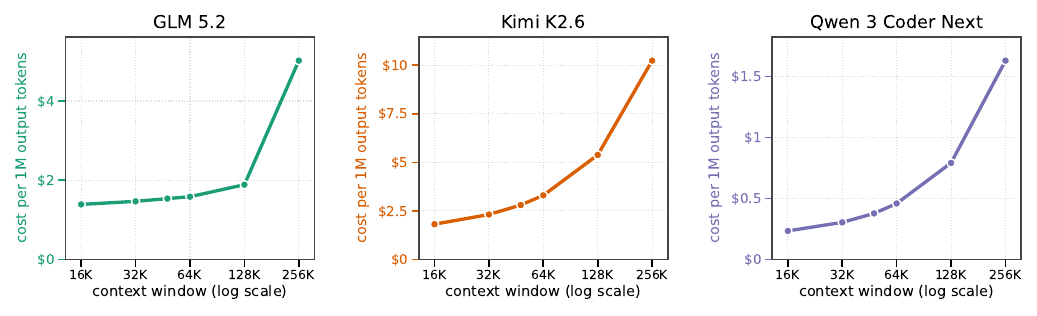}
  \caption{%
    \textbf{The cost of long context.}
    Dollar cost of generating $1$M output tokens on an existing input KV cache as a function of the context window size, obtained by converting the measured peak decode throughput of each deployment at on-demand H100 prices ($\$3.85$ per GPU-hour, Nebius, Sept.\ 2026).
    Input request lengths are drawn uniformly at random.
    GLM 5.2 and Kimi K2.6 are served with INT4 weights and BF16
    activations, Qwen 3 Coder Next is served in BF16.
    Serving at a context window size of $256$K rather than $16$K costs $3.6\times$ more per generated token for GLM, $5.6\times$ more for Kimi, and $7\times$ more for Qwen.
  }
  \label{figure:cost_per_million_tokens}
\end{figure}

Not all tokens are generated equal.
Token generation latency and memory footprint both grow with the number of tokens in a language model's context.
This is due to the inherent linear\footnote{Model architectures with a constant cost of token generation with respect to context depth have been repeatedly proposed but have never been fully embraced -- linear-cost attention-based mechanisms remain the backbone of all popular language models.} relationship between the computational cost of generating a token and its depth within a context \cite{keles2023computational}.
\Cref{figure:cost_per_million_tokens} shows the cloud token generation cost as a function of context window size for a selection of popular open-weight models.
Observe that there is a considerable monetary benefit to serving models on cloud GPUs with consistently shorter contexts.
This fact is often hidden from the end user since retail token prices are quoted at a fixed price per $1$M tokens regardless of the input context length.
Note that the extent of cost reductions ($3.6\times$ to $7\times$) rivals or exceeds that of many other orthogonal optimization methods such as aggressive quantization or structured sparsification~\cite{ashkboos2024quik,egiazarian2024aqlm,busato2020cusparselt}.

\begin{figure}[t!]
    \centering
    \includegraphics[width=0.99\linewidth]{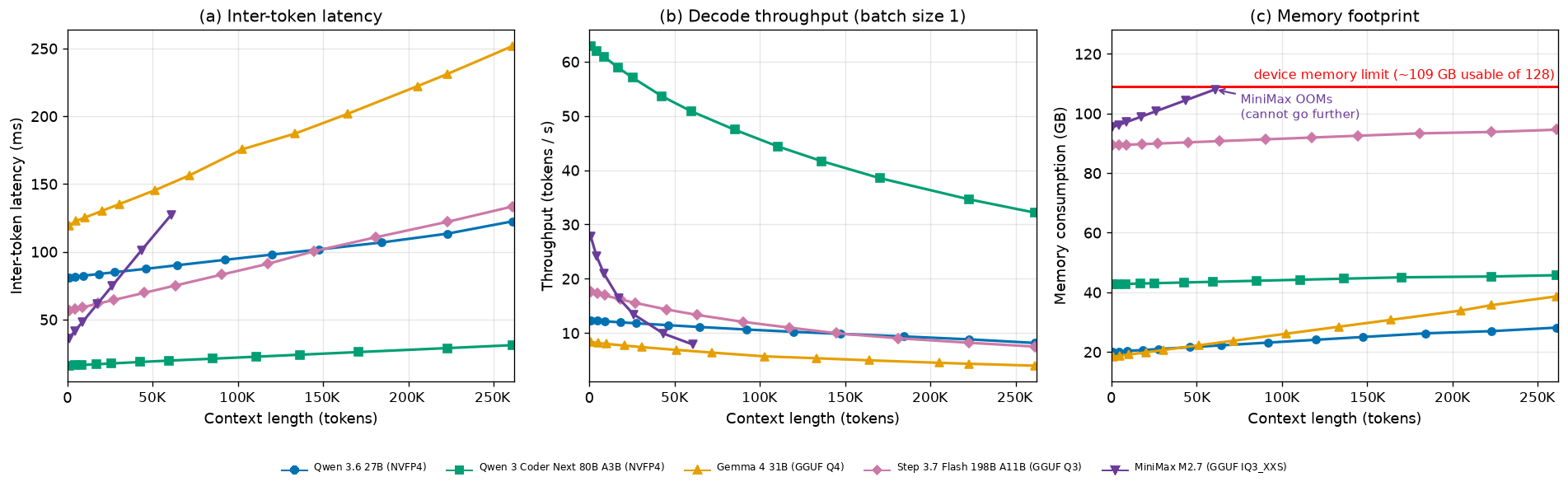}
    \caption{
      \textbf{The burden of long context.}
    The token generation latency, token throughput, and memory footprint of several open-source models measured on an NVIDIA DGX Spark plotted as a function of input context length.
    For several models, later tokens take approximately 2x longer to generate than earlier tokens and require a larger memory footprint.}
    \label{figure:models_on_spark}
    \vspace{-10pt}
\end{figure}

Monetary cost is not the only quality of LM workloads that suffers when inputs are long.
\Needspace{5\baselineskip}
\Cref{figure:models_on_spark} shows the token generation latency, throughput, and memory footprint of several popular locally-run coding models as a function of the input context length.
Observe how the generation inter-token latency approximately doubles for several models across the measured context lengths, implying a doubling of the time required to generate a response at longer contexts.

Put simply, the longer the input context provided to the language model (LM), the slower, more memory hungry, and more expensive its output token generation.

Given the considerable benefits of keeping input context short in both cloud and edge LM use, one would assume that popular LMs would have naturally evolved to operate on contexts only as short as necessary to complete their tasks.
This could not be farther from the truth.
As seen in \Cref{table:works_versus_context_window_sizes}, which compares the lengths of well-known works of writing with the context window sizes of representative proprietary and open-weight models, context window sizes have vastly outgrown many influential writings.
It would seem that models have been evolving to accommodate the ever larger possible inputs rather than to maximize efficiency.

\begin{table}[b!]
  \centering
  \small
  \setlength{\tabcolsep}{4pt}
  \renewcommand{\arraystretch}{1.08}
  \begin{tabular}{@{}p{0.319\textwidth}rr@{\hspace{1.6em}}p{0.319\textwidth}rr@{}}
    \toprule
    \multicolumn{3}{c}{\textbf{Literary works}} &
    \multicolumn{3}{c}{\textbf{Model context window sizes}} \\
    \cmidrule(lr){1-3}\cmidrule(lr){4-6}
    \textbf{Work} & \textbf{Words} & \textbf{Tokens} &
    \textbf{Model} & \textbf{Words} & \textbf{Tokens} \\
    \midrule
    The Old Man and the Sea~\cite{hemingway1952oldman} & $\sim 27$K & $\sim 37$K & MiniMax M2.7~\cite{minimax2026m27config}       & $\sim 150$K & $200$K \\
    The Great Gatsby~\cite{fitzgerald2021gatsby} & $\sim 48$K  & $\sim 64$K  & Claude Opus~4.5~\cite{anthropic2025opus45announcement}    & $\sim 150$K & $200$K \\
    Qur'an (Saheeh Int'l translation)~\cite{saheeh1997quran} & $\sim 156$K & $\sim 207$K & Gemma 4 31B~\cite{gemmateam2026gemma4}       & $\sim 192$K & $256$K \\
    King James New Testament~\cite{kingjamesbible} & $\sim 181$K & $\sim 241$K & Kimi K2.6~\cite{moonshot2026kimik26}          & $\sim 192$K & $256$K \\
    Great Expectations~\cite{dickens1998expectations} & $\sim 184$K & $\sim 245$K & Qwen 3 Coder Next~\cite{cao2026qwen3codernext}   & $\sim 197$K & $262$K \\
    The Lord of the Rings~\cite{tolkien1954lord} & $\sim 481$K & $\sim 640$K & Gemini 3.8 Flash~\cite{google2026gemini38announcement}   & $\sim 750$K & $1$M \\
    War and Peace~\cite{tolstoy2001war} & $\sim 566$K & $\sim 753$K & GLM 5.2~\cite{zai2026glm52announcement}            & $\sim 750$K & $1$M \\
    King James Old Testament~\cite{kingjamesbible} & $\sim 603$K & $\sim 802$K & GPT 5.6 Sol~\cite{openai2026gpt56}        & $\sim 790$K & $1.05$M \\
    \bottomrule
  \end{tabular}
  \caption{\textbf{Lengths of literary works against model context window sizes.} Word counts use the cited English editions. Word-token conversions use approximately $1.33$ tokens per English word, while actual counts vary with text and tokenizer. Models and their context window sizes are detailed in \cref{section:models}.}
  \label{table:works_versus_context_window_sizes}
\end{table}

The enormous context window sizes of contemporary LMs and the tangible cost advantages of keeping their contexts short thus appear to be at odds with each other.
While operating with longer contexts might in some cases be thought \textit{beneficial} to model task accuracy, we must ask:
\begin{researchquestions}[start=0]
  \item\label{rq:core_question} Are long contexts \textit{necessary}?
\end{researchquestions}
If not, let's do away with long contexts and long-context models.
And, if there are currently any objective benefits of keeping long context models, we can likely preserve them by making the appropriate post-training adjustments to future models.

To answer \cref{rq:core_question}, look at where the so-called long contexts (and, correspondingly, large context windows) in language modeling come from.
Broadly, there are two types of workloads that often result in long input contexts being passed to LMs:
\begin{longcontextworkloads}
  \item\label{workload:long_horizon} Workloads of long-horizon tasks -- tasks that may require many actions in order to be completed and where the context builds up as the model makes its way towards the completion of the task~\cite{deng2025swe,yang2024sweagent,zhou2023tianyue}.
  \item\label{workload:long_input} Workloads of long-input tasks -- tasks that require operating across a large amount of input information~\cite{bai2025longbenchv2,wang2024loong,liu2024lost}.
\end{longcontextworkloads}

Examining these two types of workloads more closely, it is not immediately clear whether operating on long contexts is necessary in order to complete them.
In particular, the entire conversational history of \cref{workload:long_horizon} does not necessarily need to be considered at every step of a long-horizon operation in order to successfully complete the task.
Likewise, the long inputs of \cref{workload:long_input} do not necessarily need to be ingested in their entirety at all times in order for the workload to conclude successfully.
Moreover, observe that the tasks in \cref{workload:long_input} can always be reduced to \cref{workload:long_horizon} by storing the long input as a resource and equipping the LM with a tool for selective input reading.

Motivated by the advantages of keeping LM generation contexts short and the apparent lack of clear justification for the popular practice for ingesting long inputs all at once and maintaining long conversational histories as discussed above, we decompose \cref{rq:core_question} to pose the following research questions:

\begin{researchquestions}
  \item\label{rq:long_horizon} Can long-horizon tasks be completed without long contexts?
  \item\label{rq:long_input} Can long-input tasks be completed without long contexts?
  \item\label{rq:single_action} Can long-horizon tasks be completed taking no more than a single action per context?
\end{researchquestions}

The rest of this paper is aimed at curtly answering these questions.

\section{Methodology}
\label{section:methodology}

We use one methodology to answer all research questions posed in \Cref{section:introduction}.

\subsection{Definitions}
\label{section:definitions}

The base term ``context'' is, unfortunately, considerably overloaded in language modelling.
Below we outline our working definitions intended to fix the meaning of otherwise vaguely interpreted terms commonly used in the community and relied on throughout this paper.
Their numerical boundaries are rough rules of thumb.

\paragraph{Inference step.}
An \emph{inference step} is one invocation of an LM to produce output.
In agentic workflows, it typically coincides with one agent turn of a conversation.

\paragraph{Input context.}
At a given inference step, the \emph{input context} is the complete sequence of tokens made available to the LM to produce output.
It may include system instructions, the task input, prior model outputs, tool results, and any other conversational history.

\paragraph{Output context.}
At a given inference step, the \emph{output context} is the complete sequence of tokens generated by the LM in the inference step.
It may include a thinking trace, tool calls, user-oriented outputs, or other model output artifacts.

\paragraph{Context.}
At a given inference step, the \emph{context} is the input context, the input context and an associated output context, or the input context and any token space reserved for a possible output context.

\paragraph{Context window and context window size.}
The \emph{context window} is the space of tokens an LM can comprehend at any given point, usually imposed by the model's architecture or indirectly by its post-training data bias.
Contexts exceeding context windows may cause system errors or unexpected behavior.
The \emph{context window size} is the maximum number of tokens that an LM endpoint can process in one invocation, whether imposed by the model's context window or narrowed by its serving configuration.

\paragraph{Long context.}
We call a context \emph{long} when its length approaches constituting a significant fraction of context window size of many contemporary frontier models, which tends to rest at approximately $256$K tokens or more at the time of this study.
We further call an execution regime or context-management strategy \emph{long-context} if it permits the context to grow toward that scale, even when an individual run terminates before reaching the context window size.

\paragraph{Short context.}
We call a context \emph{short} when its length is much closer to zero than to the context window size common to many contemporary frontier models.
In the operating regimes emphasized in this work, short contexts are commonly limited to roughly $16$K--$50$K tokens.
Note that we take ``short'' to mean short relative to common context window sizes; it does not mean that the context is small in an absolute linguistic sense.
Even context window sizes of $16$K--$50$K tokens suffice to contain elaborate literary works (cf. \Cref{table:works_versus_context_window_sizes}).

\paragraph{Long-input task.}
A \emph{long input task} is a task which is accompanied by a \emph{long input} -- i.e., input information that would constitute a long context if ingested directly.
The term \emph{long input} refers to the size of the available task input, irrespective of whether a harness places all of it in the model context at once or exposes it through selective input reading.

\paragraph{Long-horizon task.}
A \emph{long-horizon task} is a task whose completion requires many sequential model actions, with later actions depending on the observations, artifacts, or decisions produced by earlier ones.
Absent context management, the resulting interaction history accumulates across the task and may grow toward a model's context window size and form a long context.

\subsection{Benchmarks}
\label{section:benchmarks}

We assess the merits of \cref{rq:long_horizon,rq:long_input,rq:single_action} by considering two groups of benchmarks corresponding to the sources of long context identified in \Cref{section:introduction}: long-horizon tasks (\Cref{workload:long_horizon}), whose interaction histories grow as an agent acts, and long-input tasks (\Cref{workload:long_input}), whose source material can itself fill or exceed a model's context window size.
Our ensemble of benchmarks covers realistic software-engineering, information-seeking, and tool-use workloads with controlled long-input diagnostics.

\paragraph{Long-horizon task benchmarks.}
The long-horizon task benchmark group aims to represent real sequential workflows in which the model must repeatedly inspect an environment, select an action, observe its result, and use that result to decide what to do next.
Without context management, every command, tool response, retrieved document, and intermediate decision is appended to an ever-growing history.
The software-engineering benchmarks are particularly representative examples because repository exploration, implementation, and validation naturally form long sequences of interdependent actions.
The remaining benchmarks broaden this group to information retrieval, user interaction, and cross-tool orchestration.

\begin{itemize}
    \item \textbf{SWE-bench Verified} \cite{jimenez2024swe} is a human-validated set of 500 real GitHub issue-resolution tasks from popular repositories, where the agent must produce a patch that passes the repository's held-out test suite.

    \item \textbf{SWE-bench Multilingual} \cite{yang2025swesmith} comprises 300 manually curated issue-resolution tasks from repositories across nine programming languages.
    It tests repository exploration, implementation, and validation across different language ecosystems.
    
    \item \textbf{Terminal-Bench 2.1} \cite{merrill2026terminal} is an agentic benchmark of 89 real command-line tasks executed in sandboxed containers and scored by per-task verifiers, measuring end-to-end terminal problem-solving beyond single-repository patching.

    \item \textbf{FRAMES} \cite{krishna2025frames} contains 824 challenging multi-hop questions whose answers require combining information from two to fifteen Wikipedia articles.
    In our agentic formulation, the model searches a fixed Wikipedia corpus, opens relevant articles incrementally, and submits a free-form answer evaluated against the reference answer by an LM judge.
    The benchmark consequently tests iterative retrieval and evidence integration while allowing the retrieved evidence to accumulate in the interaction history.

    \item \textbf{$\tau^2$-bench} \cite{barres2025tau2bench} evaluates agents that converse with a simulated user while operating domain-specific tools under explicit policy constraints.
    We use the 184-task golden set spanning retail, airline, and telecommunications; success requires both satisfying the user's request and bringing the environment to the correct final state.
    Its mixture of dialogue, reasoning, and state-changing tool calls tests whether information established early in an interaction remains usable by later actions.

    \item \textbf{BFCLv3} \cite{patil2025berkeley} evaluates multi-turn tool use in simulated environments.
    We use its 200 long-context episodes, which require agents to retain information across user turns and schedule state-changing tool calls.
\end{itemize}

\paragraph{Long-input task benchmarks.}
The long-input task benchmark group holds the task question comparatively short while making the
information needed to answer it large, at least in principle.
It includes three broad, realistic long-document QA benchmarks and four needle-style benchmarks.

\begin{itemize}
    \item \textbf{LongBench v2} \cite{bai2025longbenchv2} comprises 503 challenging multiple-choice questions with heterogeneous, predominantly natural inputs ranging from $8$K to approximately $2$M words.
    Its six task families cover single- and multi-document question answering, long in-context learning, long dialogue, code-repository understanding, and structured-data understanding.

    \item \textbf{LOONG} \cite{wang2024loong} is a 1,600-question multi-document benchmark in which a question typically depends on multiple complete financial reports, legal cases, or academic papers in either English or Chinese.
    Its locating, comparison, clustering, and chain-of-reasoning tasks are designed so that evidence is distributed across multiple load-bearing documents.

    \item \textbf{Artificial Analysis Long Context Reasoning (AA-LCR)} \cite{artificialanalysis2025lcr} comprises 100 human-crafted questions requiring reasoning across sets of real-world documents averaging approximately $100$K input tokens.
    Its seven document categories include company and industry reports, government consultations, academic and legal documents, marketing materials, and survey reports.
    Answers require integrating evidence and performing multi-step reasoning across documents, rather than retrieving a directly stated fact.

    \item \textbf{Needle in a Haystack (NIAH)} \cite{kamradt2023needle} consists of target sentences inserted at controlled depths in long bodies of unrelated essay text and of asking for the fact expressed by that sentence.
    Our 125-instance grid varies both insertion depth and document length, with contexts extending from $16$K to roughly $500$K tokens.

    \item \textbf{RULER} \cite{hsieh2024ruler} is a synthetic long-context suite containing thirteen tasks: eight needle-retrieval variants, variable tracking, common- and frequent-word extraction, and two multi-hop question-answering tasks.
    We evaluate all thirteen at $256$K and $500$K tokens and score the recall of the required answer set.

    \item \textbf{Counting-Stars} \cite{song2024countingstars} scatters 32 marked numbers throughout a long text and asks the model to recover them all.
    We use the complete 192-question public set, spanning English and Chinese, acquisition and reasoning variants, and documents from $4$K to $128$K tokens.
    The reasoning variant includes initially incorrect counts that are subsequently corrected, requiring the model to distinguish the final value from a nearby distractor rather than merely copy every marked number.

    \item \textbf{BABILong} \cite{kuratov2024babilong} consists of short facts from the bAbI \cite{weston2015babi} reasoning tasks embedded into book text.
    Answering requires locating the relevant statements and then performing operations such as fact chaining, counting, negation, or set tracking.
    We evaluate 550 questions across all ten task types and five input lengths from $64$K to $1$M tokens.
\end{itemize}

Note that even the needle-style benchmarks considered often do not feature a clear lexical overlap aiding easy retrieval.
They do, however, require aggregation, variable tracking, or reasoning over multiple retrieved facts for the correct answer to be produced.

\subsection{Models}
\label{section:models}

We perform our study on two groups of models.
Large closed-weight models are used as the workers in our principal task-accuracy experiments, while large open-weight models support the cloud system throughput study and complementary local experiments.

\paragraph{Large closed-weight models.}
Our principal benchmark experiments use \textbf{Claude Opus~4.5} and \textbf{GPT 5.6 Sol}.
We select these models as representative of the strong general-purpose models used for coding, tool use, and other long-horizon workflows.
Using independently developed model families also reduces the risk that an observed effect of compaction is specific to one provider or post-training recipe.

\paragraph{Large open-weight models.}
We use large open-weight models for two additional parts of the experimentation.
First, the cloud-throughput study in \Cref{figure:cost_per_million_tokens} uses models large enough to require multi-GPU deployment and exposes the effect of context window size under saturated serving load:

\begin{itemize}
    \item \textbf{GLM 5.2} \cite{glm5team2026glm5,zai2026glm52}, evaluated at context window sizes up to $1$M tokens;
    \item \textbf{Kimi K2.6} \cite{moonshot2026kimik26}, evaluated at context window sizes up to $256$K tokens; and
    \item \textbf{Qwen 3 Coder Next} \cite{cao2026qwen3codernext}, evaluated at context window sizes up to $256$K tokens.
\end{itemize}

We fit the context-cost ratio from these throughput measurements and define the \Cref{rq:long_horizon} optimized-retail-cost estimate in \Cref{appendix:retail_cost_fit}.

Second, we use locally deployable open-weight models for the on-device systems measurements in \Cref{figure:models_on_spark} and for complementary task-level experiments.
We deploy them on an NVIDIA DGX Spark \cite{nvidia2025dgxspark}, whose 128GB of unified memory leaves about 109GB available for local LM serving.

Such deployments are relevant when source code or other task data must remain on-device for privacy or cost reasons.
The two Qwen checkpoints are served in NVFP4 \cite{alvarez2025nvfp4}; the remaining three are served from GGUF weights through \texttt{llama.cpp} using 3- to 4-bit \emph{k}-/\emph{i}-quantization.
Together they cover hybrid linear/softmax attention, distinct interleaved sliding-window/global-attention stacks, gated grouped-query attention, and conventional full grouped-query softmax attention.
This architectural diversity helps to test the sufficiency of short context claimed in \Cref{rq:long_horizon} across models with different attention mechanisms and post-training characteristics.

\begin{itemize}
    \item \textbf{Qwen 3.6 27B} \cite{qwen2026qwen36} is a 27B-parameter dense model served in NVFP4.
    Its layers alternate in a 3:1 pattern between Gated DeltaNet and gated full attention.
    It has a context window size of $256$K tokens.
    
    \item \textbf{Qwen 3 Coder Next} \cite{cao2026qwen3codernext} is a sparse MoE model with 80B total and 3B active parameters, served in NVFP4.
    It uses the same hybrid Gated DeltaNet/full-attention pattern and has a context window size of $256$K tokens.

    \item \textbf{Gemma 4 31B} \cite{gemmateam2026gemma4} is a 31B-parameter dense model served from GGUF \texttt{Q4\_K\_M} weights.
    It interleaves $1024$-token local sliding-window attention with global-attention layers and has a context window size of $256$K tokens.

    \item \textbf{Step 3.7 Flash (198B A11B)} \cite{stepfun2026step37flash} is a sparse MoE model with 198B total and 11B active parameters, served from GGUF \texttt{Q3\_K\_M} weights.
    It interleaves $512$-token sliding-window and full-attention layers in a 3:1 pattern, uses head-wise gated grouped-query attention, and has a context window size of $256$K tokens.

    \item \textbf{MiniMax M2.7} \cite{chen2026minimaxm2} is a sparse MoE model with 230B total and 10B active parameters, served from GGUF \texttt{IQ3\_XXS} weights.
    It uses full-softmax grouped-query attention throughout and has a context window size of $200$K tokens~\cite{minimax2026m27config}.
\end{itemize}

\subsection{Harnesses}
\label{section:harnesses}

We use two closely related base harnesses.
Both implement a minimal ReAct loop: the task is supplied in an initial user message, the model issues one Bash action at a time, and each action and its result are appended to the conversation \cite{yao2023react}.
When used, the Bash environment exposes the files, programs, and benchmark-specific utilities available to the agent.
Each Bash invocation treats the related standard output, return code, and any execution error as action results.

\paragraph{\texttt{mini-swe-agent}.}
For software-engineering and terminal tasks, we use the released \texttt{mini-swe-agent} harness \cite{minisweagent}, with only the environment integration required by each benchmark.
Its system prompt describes the software-engineering workflow, and its Bash environment provides the task repository or container.
When done, the agent submits its proposed pull request as a patch for evaluation.

\paragraph{\texttt{mini-agent}.}
For the remaining benchmarks, we use our \texttt{mini-agent} harness, which is a straightforward generalization of \texttt{mini-swe-agent} away from software engineering.
It retains the same agent loop, model interface, and Bash interface, but uses benchmark-specific instructions, environments, and submission rules.
The instructions specify the available commands, task procedure, and required answer format.
Only the system instruction and task description are provided in the initial user message.
Supporting inputs and tools are made available through the environment.
When done, an evaluation benchmark adapter recognizes the designated submission command, extracts its payload, terminates the episode, and records the prediction in the benchmark evaluator's required format.

\subsubsection{Accommodations for long conversational context}
\label{section:long_conversations}

\begin{figure}[H]
    \centering
    \includegraphics[width=0.99\linewidth]{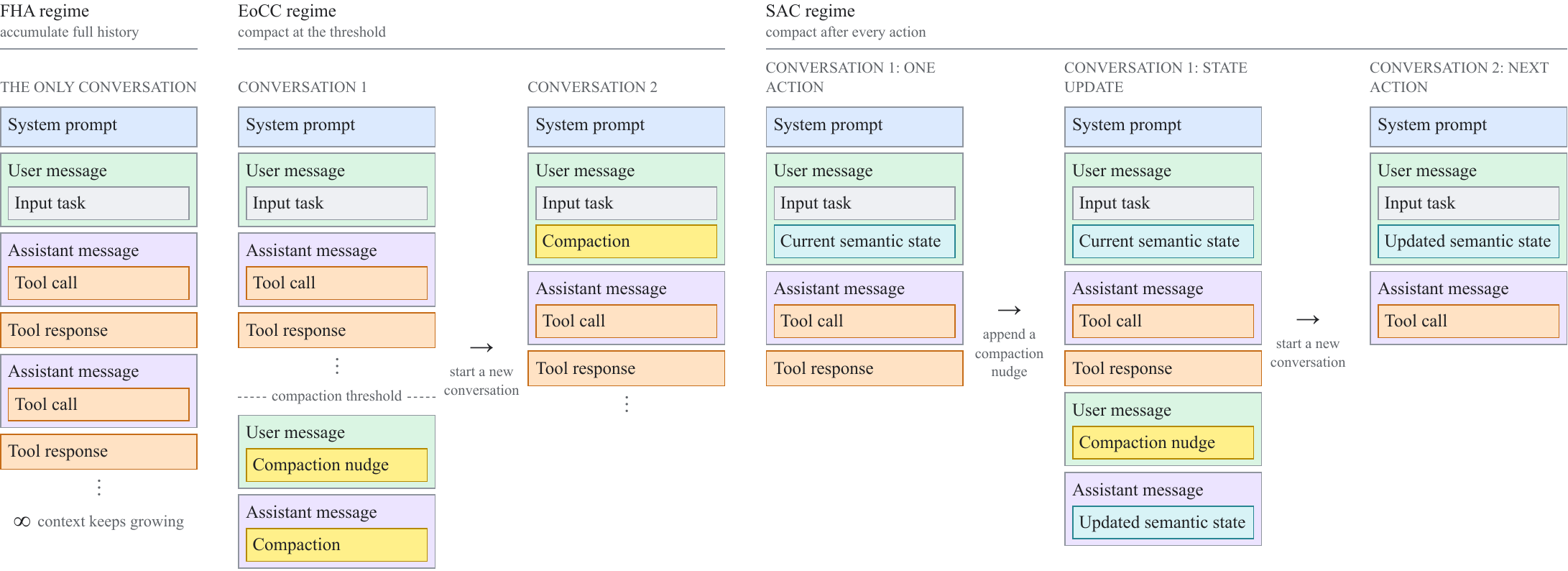}
    \caption{\textbf{Context-management regimes.}
    The three context-management regimes considered in this work: Full-History Accumulation (FHA), End-of-Conversation Compaction (EoCC), and Single-Action Compaction (SAC).}
    \label{figure:compaction_regimes}
\end{figure}

We compare three context-management regimes applied to the base harnesses described above.
These regimes determine which interaction history is presented to the model.
The regimes' functioning is illustrated in \Cref{figure:compaction_regimes}.

\paragraph{Full-History Accumulation (FHA).}
The baseline full-history accumulation regime appends every model action and environment observation to a single conversation for the duration of the task.
No history is removed or replaced during a run.
This is the default mode of operation for \texttt{mini-swe-agent} and \texttt{mini-agent}.

\paragraph{End-of-Conversation Compaction (EoCC).}
This regime follows the full-history ReAct loop until the length of the conversational context crosses a pre-specified compaction threshold.
The model is then instructed within the active conversation to produce a bounded, task-relevant compaction of its state (see \Cref{appendix:good_compaction}).
A higher total context limit reserves space for this compaction turn and terminates the conversation if compaction does not complete in time.
The harness starts a fresh conversation containing the original task and the compacted state, and execution resumes under the same system prompt.
This cycle repeats whenever the compaction threshold is crossed.
We write these two values as ``total context limit / compaction threshold'' in tokens.

\paragraph{Single-Action Compaction (SAC).}
This regime maintains a compact semantic state and essentially no conversational history.
Each conversation receives the task and current semantic state, performs exactly one environment action, observes its result, and updates the semantic state.
The harness then discards that conversation and starts a fresh one for the next action.
For the experiments in \Cref{section:single_action_answer}, we apply SAC as a context-management regime to \texttt{mini-swe-agent}; the underlying agent harness, model, and task environment otherwise remain unchanged.
SAC therefore keeps the active conversational context shortest of the three regimes and operationalizes the single-action setting of \cref{rq:single_action}.

\paragraph{Context length measures.}
We distinguish \emph{total length} from \emph{peak length}.
For a full-history baseline, total length is the length of its execution's conversation.
For EoCC and SAC, total length is the sum of the final lengths of all conversations created during the execution.
Peak length is the length of the largest conversation produced during the task.
Both measures are expressed in thousands of tokens (K) and summarized across tasks as mean\,/\,median\,/\,maximum or median\,/\,maximum, as indicated in each table.

\subsubsection{Accommodations for long inputs}
\label{section:long_inputs}

We enable selective input reading by storing each long input as one or more documents (e.g., \texttt{input01.txt}, \texttt{input02.txt}) in the agent's filesystem environment rather than placing their full contents in the initial conversation.
The task message identifies the available documents, and the agent performs selective input reading using standard Bash tools (e.g., \texttt{sed}, \texttt{awk}, \texttt{head}).
This method of input presentation makes selective input reading an explicit environment action and thereby reduces a long-input workload to a long-horizon workload.
We provide the agent instructions and examples of selective reading in \Cref{appendix:selective_reading}.

\subsection{Latency and throughput measurement}
\label{section:latency_and_throughput_measurement}

\begin{figure*}[t]
  \centering
  \includegraphics[width=\textwidth]{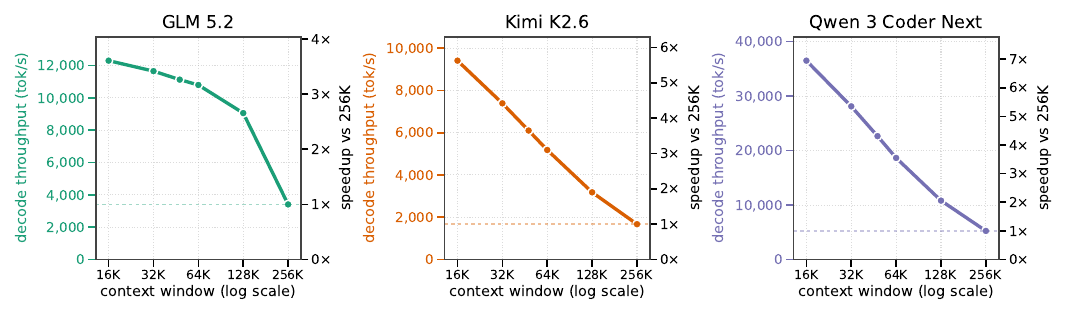}
  \caption{%
    \textbf{Decode throughput collapses as the context window size grows.}
    Peak sustained decode throughput (tokens/s, left axis) for open-weight coding models served under vLLM on NVIDIA H100 GPUs, as a function of the configured context window size.
    GLM 5.2 and Kimi K2.6 are served with INT4 experts and BF16 activations (Kimi additionally with an FP8 KV cache), Qwen 3 Coder Next in BF16.
    The right axis re-expresses the same curve as a speedup over the $256$K operating point.
    Reducing the context window size from $256$K to $16$K buys $3.6\times$ (GLM 5.2), $5.6\times$ (Kimi K2.6), and $7\times$ (Qwen 3 Coder Next) more decoded tokens per second on identical hardware.
  }
  \label{figure:throughput_by_context}
\end{figure*}

\paragraph{Cloud serving.}
For each model, we use the smallest H100 node-deployment that can hold the served checkpoint and a KV cache at the maximum evaluated context window size: eight H100 GPUs for Qwen 3 Coder Next and sixteen for GLM~5.2 and Kimi~K2.6.
For each context window size, we draw request lengths uniformly from nested prefixes of a shared token sequence, with sufficient space reserved for generation.
We prefill the longest prefix once, enable prefix caching, and require every subsequent prefix request to generate exactly $1,024$ tokens.
After warm-up, we sweep request concurrency and measure sustained aggregate decode throughput as completed output tokens divided by elapsed time.
We report the highest-throughput setting with no KV-cache preemption; repeated measurements varied by at most 2\%.
The results are in \Cref{figure:throughput_by_context}.
The cost values in \Cref{figure:cost_per_million_tokens} convert these throughput measurements to the cost of $1$M output tokens using the stated hourly GPU price.

\paragraph{Local serving.}
On the DGX Spark, we evaluate one model instance and one request at a time.
For each model, we construct a sequence at its maximum supported length and measure generation after prefixes of increasing length drawn from that sequence.
At each length, we record the mean inter-token latency over the next $1,024$ generated tokens and the total device-memory consumption.
The batch-one decode throughput in \Cref{figure:models_on_spark} is the reciprocal of the mean inter-token latency.

\clearpage
\Needspace{12\baselineskip}
\section{Long-horizon tasks don't need long contexts}
\label{section:long_horizon_answer}

\rqsectionbox{rq:long_horizon}
  {Can long-horizon tasks be completed without long contexts?}
  {Yes. Frequent compaction keeps the context short while broadly maintaining task accuracy.}

Following the outline of \Cref{section:methodology}, we evaluate EoCC on 6 benchmarks representing long-horizon workloads and measure the performance of EoCC-modified harnesses against their baseline FHA counterparts.
For each benchmark and model, we sequentially decrease the total context token limit and the corresponding compaction threshold to fit the total context limit.
This has the effect of lowering the maximum context window needed by the model and of triggering increasingly more frequent compactions while executing long-horizon tasks.
We describe what these compactions are designed to preserve in \Cref{appendix:good_compaction}.
For SWE-bench Verified, TerminalBench 2.1, SWE-bench Multilingual, and FRAMES, we evaluate the models at total context limits / compaction thresholds (cf. \Cref{section:long_conversations}) of $48$K/$36$K, $32$K/$24$K, $24$K/$18$K, and $16$K/$12$K.
For BFCLv3, we consider $24$K/$18$K, $16$K/$12$K, and $12$K/$8$K, while for Tau2Bench, we consider $16$K/$12$K, $12$K/$8$K, and $8$K/$6$K.

\begin{figure}[H]
\centering
\includegraphics[width=\textwidth]{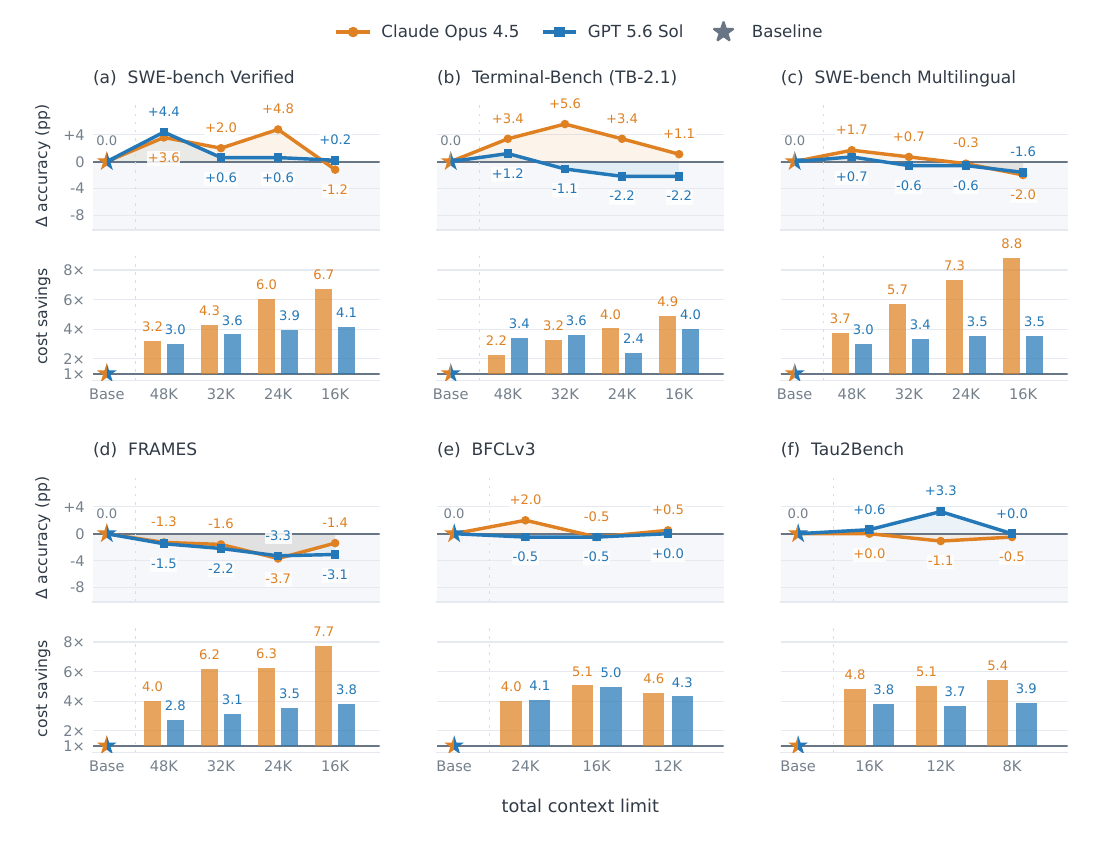}
\caption{
  \textbf{Long-horizon results as a function of total context limit.}
  Top: accuracy changes in percentage points relative to each model's full-history baseline; zero indicates unchanged accuracy.
  Bottom: baseline estimated optimized retail cost divided by the compaction configuration's estimated optimized retail cost; $1\times$ indicates unchanged cost.
  Orange/blue identify Claude Opus~4.5 and GPT 5.6~Sol, respectively, and stars mark the baselines.
  Cost estimates use the model in \Cref{appendix:retail_cost_fit}.
}
\label{figure:long_horizon_results}
\end{figure}

\Cref{figure:long_horizon_results} plots the EoCC accuracies and estimated cost savings relative to FHA baselines against increasing levels of compaction.

\vfill

Skimming through the line plots, observe first that the performance broadly remains flat, sometimes falling a few points below, sometimes remaining elevated above the baseline.
More specifically, across all benchmarks and compaction thresholds, the relative performance of Opus 4.5 ranges from -3.7 to +5.6ppts, while the relative performance of GPT 5.6 Sol ranges from -3.3 to +4.4.
The ranges are somewhat deceiving when considered by themselves, however: for either model, no more than 5 out of 22 threshold configurations lose more than 2ppts of the baseline.
Secondly, observe the multiples of cost savings across all benchmarks, ranging from 2.2x to 8.8x for Opus 4.5 and from 2.4x to 5.0x for GPT 5.6 Sol.
These represent the estimated cost savings as a consequence of serving only shorter context window sizes, and are considerable.

One pathology that could occur due to frequent compaction is that while it may successfully enforce smaller context window size, the full workflow history might be becoming longer as a consequence of the model having to re-do previously done tasks or to revisit previously seen information.
Such behavior could result in an increased total token spend per task, ultimately erasing some of the cost savings due to smaller context windows. 
In our experimentation, this was generally not the case.
Detailed per-benchmark accuracy, token spend, total length, cost and their discussion are provided in \Cref{appendix:long_horizon_results}.

In sum, we observe that even the naive approach to keeping context short by compacting agent's work history once its conversation reaches a certain restrictive threshold leads to, broadly-speaking, on-par performance in model performance on long-horizon tasks when using frontier closed-weight models.
Complementary results for open-weight models are provided in \Cref{appendix:open_weight_results}.
It would seem that the benefits of maintaining long contexts manifest themselves inconsistently and are not drastic so as to discourage efforts to keep the working context short in autonomous agentic workloads.
Hence our answer to \Cref{rq:long_horizon}: yes.

\subsection{Long-horizon tasks may sometimes need the history}
Long-horizon tasks may sometimes need the history, but that does not mean that they require long contexts.

Compaction is a lossy context compression method.
Information will be lost as a consequence of compacting a conversational history.
Even more information will be lost if the conversational history being compacted already contains a result of an earlier compaction.

It is not too difficult to construct a task whose accuracy is guaranteed to suffer when a compaction is performed.
Consider, for example, an agent that has access to a tool that generates coin tosses and reports the outcome of each coin toss in its tool result.
With a compaction budget fixed, the agent is guaranteed to eventually start losing information about the outcomes of some of the coin tosses.
If, at some point, the agent needs to inspect the sequence of coin toss outcomes (e.g., access the $i$th outcome), it is guaranteed to eventually fail as a consequence of some compaction necessarily eliminating some of its toss outcome records, with the compaction being the cause.

Observe, however, that any such example can be eliminated by creating a harness that keeps a record of every message of any agentic conversation, and makes it readily accessible to the agent by the means of a selective reading tool.
The querying of the agent's conversational history then becomes a long-input task, which we tackle in \Cref{rq:long_input}.

In other words, there is a difference between information that needs to be a part of model context and information that could, in principle, be retrieved selectively on a ``need-to-know basis''.

Note that the findings of our experimentation above are not that ``you don't ever need to access conversational history'', but roughly that ``you don't need to keep the entire conversational history in the context in order to complete long-horizon tasks''.

\clearpage
\Needspace{12\baselineskip}
\section{Long-input tasks don't need long contexts}
\label{section:long_input_answer}

\rqsectionbox{rq:long_input}
  {Can long-input tasks be completed without long contexts?}
  {Yes. Selective input reading and compaction keep contexts short, maintaining task accuracy.}

We evaluate EoCC on 7 benchmarks representing long-input workloads and measure the performance of EoCC-modified harnesses against their baseline full-input counterparts.
For each benchmark and model, the full-input baseline receives the entirety of the input as prescribed by the benchmark while the compacting harness uses Bash tools to perform selective reading of inputs from environment filesystem (see \Cref{appendix:selective_reading}).
Where appropriate in EoCC runs, we further sequentially decrease the total context token limit and the corresponding compaction threshold to fit the total context limit.
For many long-input tasks (especially the needle-based tasks), one restrictive total token limit is enough to showcase on-par or better performance.

\begin{figure}[H]
\centering
\includegraphics[width=0.82\textwidth]{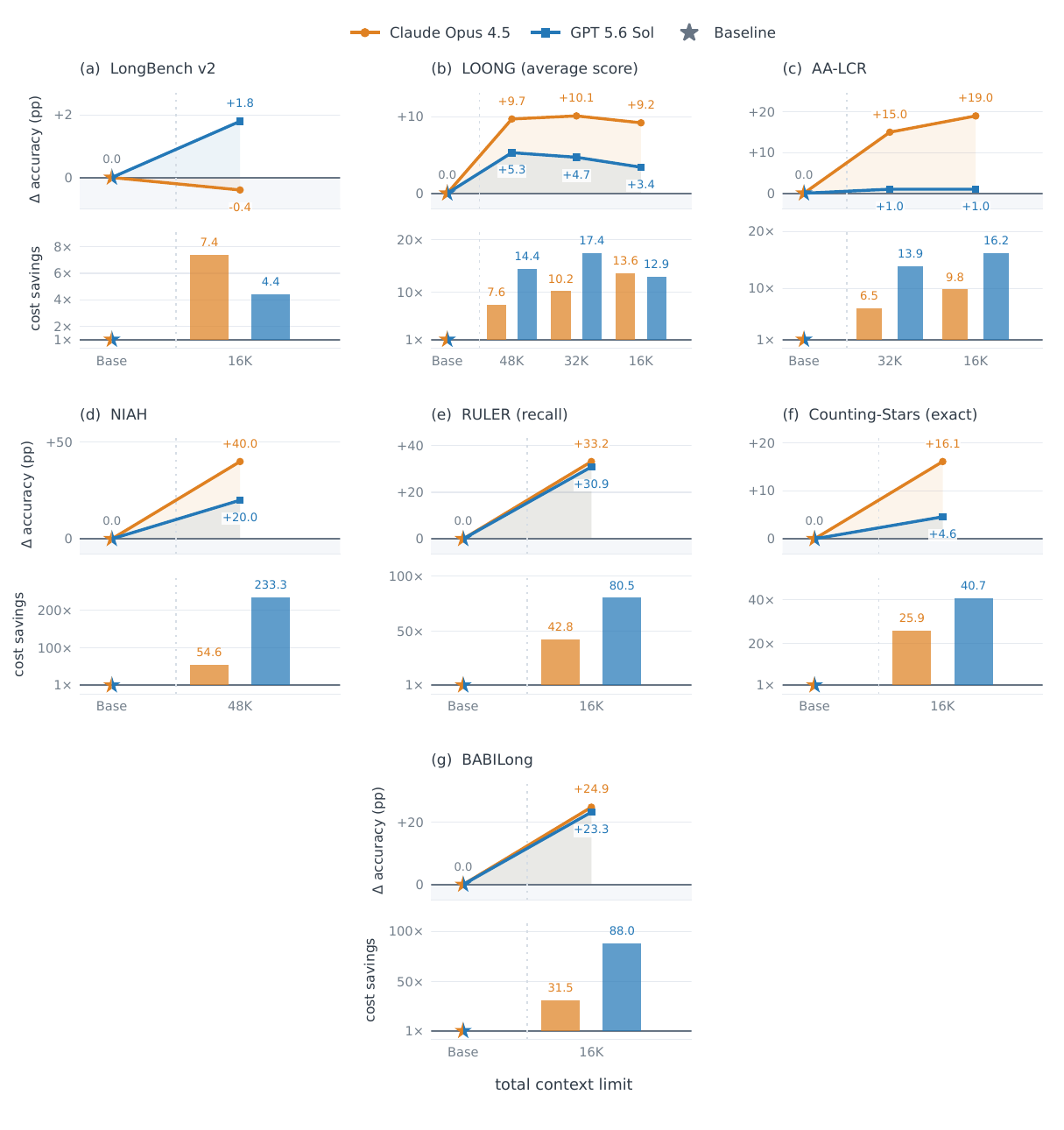}
\caption{
  \textbf{Long-input results as a function of total context limit.}
  Top: accuracy changes relative to each model's full-input baseline.
  Bottom: optimized retail cost savings, using the model in \Cref{appendix:retail_cost_fit}.
}
\label{figure:long_input_results}
\end{figure}

\Cref{figure:long_input_results} plots the EoCC accuracies and estimated cost savings relative to FHA baselines against one or multiple levels of compaction.

\vfill

Looking through the line plots, observe first that the performance broadly remains flat, sometimes falling a few points below, sometimes remaining elevated above the baseline.
More specifically, across all benchmarks and compaction thresholds, the relative performance of Opus 4.5 ranges from -0.4 to +40.0ppts, while the relative performance of GPT 5.6 Sol ranges from +1.0 to +30.9 (!).
Observe also the multiples of cost savings across all benchmarks, ranging from 6.5x to 54.6x for Opus 4.5 and from 4.4x to 233.3x for GPT 5.6 Sol.
These represent the estimated cost savings as a consequence of both selective input reading and serving only shorter context window sizes, and warrant serious consideration.

\begin{figure}[H]
\centering
\includegraphics[width=0.82\textwidth]{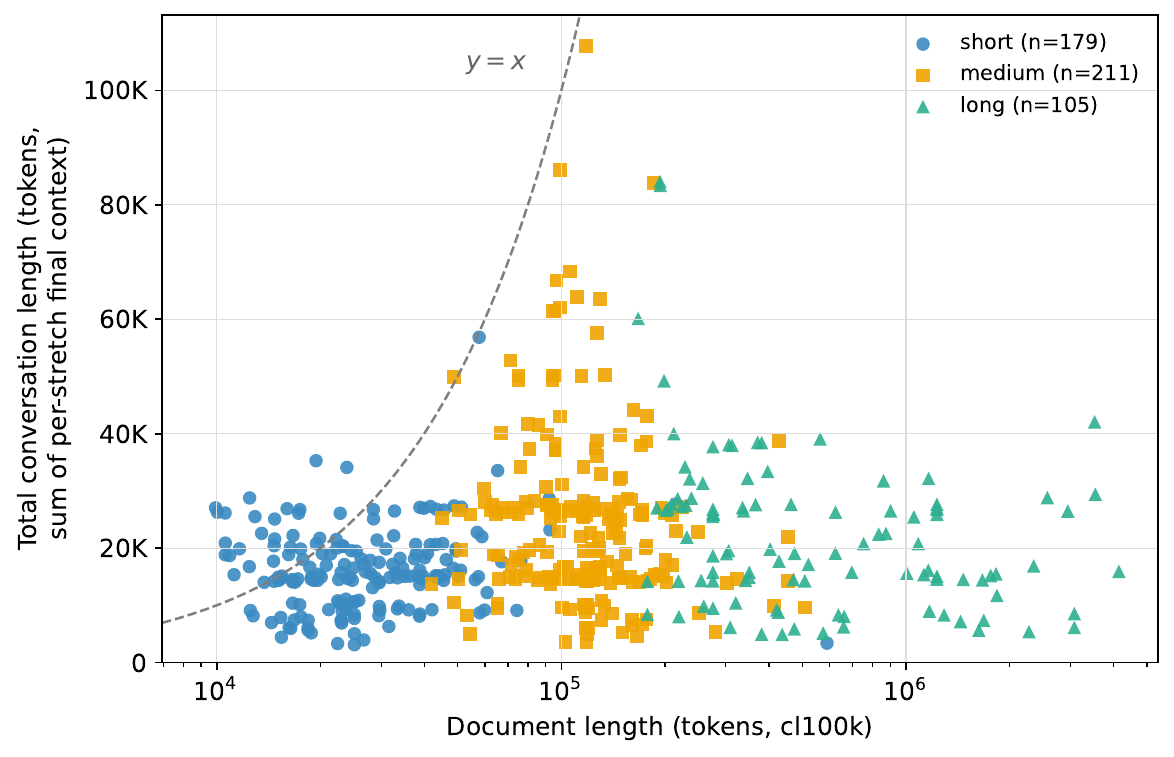}
\caption{\textbf{LongBench v2: conversation length versus document length.}
Claude Opus~4.5, compaction at $16$K / $12$K.
The document-length axis is logarithmic and the conversation-length axis is linear.
The dashed curve marks $y=x$.
}
\label{figure:longbench_context_scatter}
\end{figure}

\begin{figure}[H]
\centering
\includegraphics[width=0.82\textwidth]{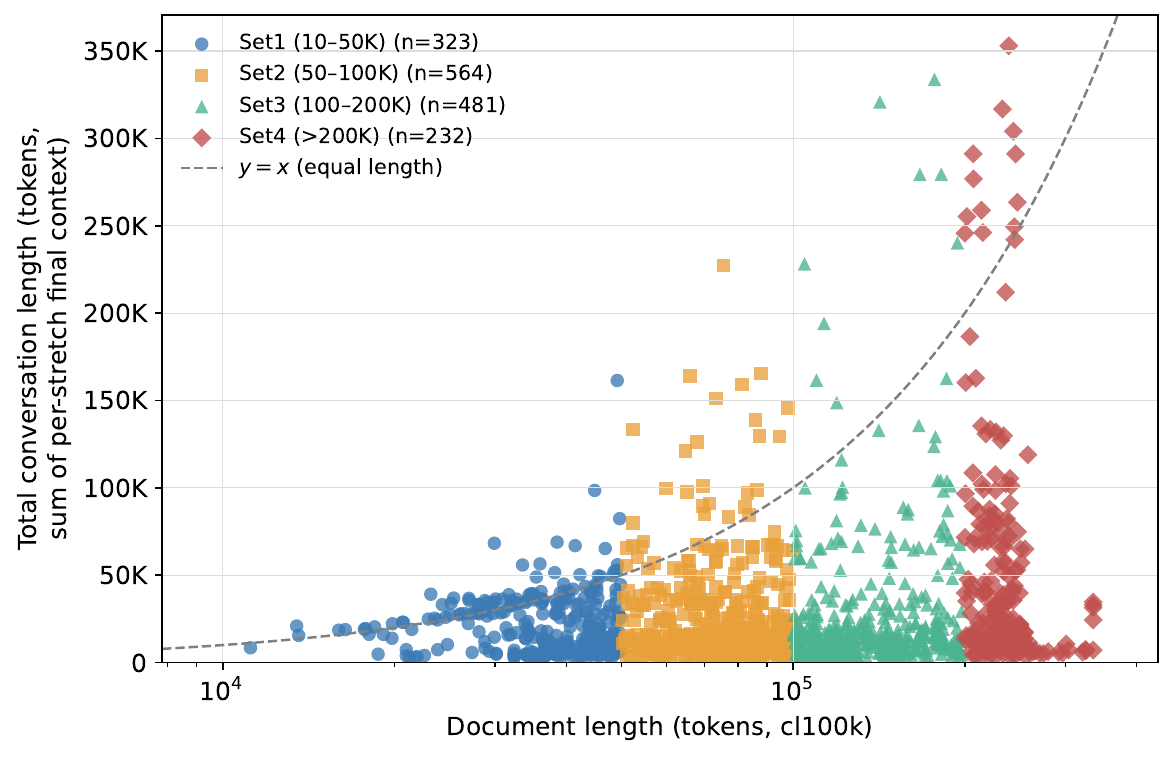}
\caption{\textbf{LOONG: conversation length versus document length.} Claude Opus~4.5, compaction at $48$K / $40$K (1,600 questions).
The document-length axis is logarithmic and the conversation-length axis is linear.
The dashed curve marks $y=x$.
}
\label{figure:loong_context_scatter}
\end{figure}

\Cref{figure:longbench_context_scatter,figure:loong_context_scatter} plot total conversation length, summed across stretches, against document length.
On logarithmic document-length axes spanning orders of magnitude, conversation lengths on the linear vertical axes remain concentrated in broadly similar bands for most tasks, with some longer outliers.
The dashed equal-length reference $y=x$ rises exponentially in these plotting coordinates, whereas most conversations do not grow proportionally with document length.
This shows that even as document length increases, the conversation length remains broadly bounded.

Detailed per-benchmark accuracy, token spend, cost, compactions and their discussion are provided in \Cref{appendix:long_input_results}.

In summary, we observe that selective input reading, with compaction available as a fallback, matches or improves long input baseline performance across almost all long-input benchmarks and remains close to baseline in the remaining case, while keeping peak live context far below baseline conversation lengths.
The near-zero median number of compactions and the roughly constant conversation lengths as document size grows support the intuition that despite the long inputs provided to tasks, the actual amount of information required to solve those tasks remains low.
Thus, the size of a task's input does not itself establish a need for long context.
Hence our answer to \Cref{rq:long_input}: yes.

\Needspace{32\baselineskip}
\section{Compaction can be performed after every action}
\label{section:single_action_answer}

\rqsectionbox{rq:single_action}
  {Can long-horizon tasks be completed taking no more than a single action per context?}
  {Yes. Single-action compaction maintains an effective compact semantic state.}

The combined comparison in \Cref{table:sac_profile} shows that, for Claude Opus~4.5, SAC preserves the \texttt{mini-swe-agent} baseline accuracy on SWE-bench Verified and improves it by 4.6 percentage points on Terminal-Bench.
For GPT 5.6 Sol, SAC reaches 82.4\% on SWE-bench Verified and 84.3\% on Terminal-Bench (75 of 89 tasks solved), below its mini baseline by 2.0 and 3.3 percentage points, respectively.
Meanwhile, across both models and benchmarks, median peak live context under SAC is only $7$--$9$K tokens, and the maximum falls from $130$--$200$K to $15$--$37$K tokens.

\begin{table}[H]
\centering
\captiontable{%
\resizebox{\textwidth}{!}{%
\begin{tabular}{lcccccc}
\toprule
 & \multicolumn{3}{c}{Accuracy (\%)} & \multicolumn{3}{c}{Peak context length (K)} \\
\cmidrule(lr){2-4}\cmidrule(lr){5-7}
Benchmark & mini & Best EoCC & SAC & mini & Best EoCC & SAC \\
 & baseline & & & (median\,/\,max) & (median\,/\,max) & (median\,/\,max) \\
\midrule
\multicolumn{7}{l}{\textbf{Claude Opus~4.5}} \\
SWE-bench Verified   & 70.0 & 74.8 \; ($24$K / $18$K) & 70.2 & 32\,/\,200 & 19\,/\,22 & 8\,/\,16 \\
Terminal-Bench (2.1) & 47.2 & 52.8 \; ($32$K / $24$K) & 51.8 & 27\,/\,130 & 24\,/\,31 & 9\,/\,19 \\
\midrule
\multicolumn{7}{l}{\textbf{GPT 5.6 Sol}} \\
SWE-bench Verified   & 84.4 & 88.8 \; ($48$K / $36$K) & 82.4 & 18\,/\,173 & 22\,/\,35 & 7\,/\,15 \\
Terminal-Bench (2.1) & 87.6 & 88.8 \; ($48$K / $36$K) & 84.3 & 17\,/\,133 & 23\,/\,40 & 8\,/\,37 \\
\bottomrule
\end{tabular}}}{\textbf{SAC versus FHA and EoCC.}
The \emph{mini baseline} is the full-history accumulation regime.
The \emph{Best EoCC} accuracy is accompanied by its total context limit /
compaction threshold in parentheses.
The \emph{Best EoCC} and \emph{SAC} columns under \emph{Peak context length} report the largest single-call contexts under the best EoCC configuration and SAC, respectively; all context statistics are per task in K tokens and reported as median\,/\,maximum.
\label{table:sac_profile}}
\end{table}

Thus, SAC preserves or improves baseline accuracy for Claude Opus~4.5, while GPT 5.6 Sol incurs small accuracy reductions.
This is somewhat remarkable, considering that both models reduce each interaction to a single action and carry only a compact semantic state between contexts.
The resulting live contexts are far shorter than the full-history conversations, showing that compaction frequency can be pushed to relative extremes without catastrophic consequences for task accuracy.
Hence our answer to \Cref{rq:single_action}: yes.

\subsection{Semantic State Space Models (SSSMs).}
The above observations of \Cref{section:single_action_answer} and earlier observations of \Cref{section:long_horizon_answer,section:long_input_answer} force a reflection: If powerful models can broadly maintain performance by compacting often or even compacting after taking a single action, cannot all long-context workflows be coerced into becoming single-action workflows through compaction?
In other words, could not all models potentially be performing single-action compaction (SAC) by default?
To do so would roughly correspond to maintaining a \textbf{semantic state} across long-context workflows.

We propose the notion of a Semantic State Space Model (SSSM).
A SSSM is a model post-trained to (1) comprehend an input language description of a state of a conversation, (2) reason about the appropriate next action or about the task at hand, (3) take an action, (4) register its outcome, and (5) update the language description of the conversation's state with the reasoning conclusions, action decisions, and action outcomes in a way that allows a separate inference instance of the SSSM to continue the conversation toward completion.
The notion of a SSSM is closely related to that of a State-Space Model \cite{NEURIPS2021_05546b0e,gu2022efficientlymodelinglongsequences,gu2024mambalineartimesequencemodeling}.
Unlike an SSM, which maintains a black-box neural state between token generations, an SSSM maintains an explainable natural language (hence ``semantic'') state between turns of a continuing conversation.

The utility of SSSMs as opposed to existing conversational models trained for growing full-history conversations lies in their increased token efficiency -- as shown in \Cref{figure:cost_per_million_tokens}, there is a considerable cost benefit to maintaining short contexts.
We further surmise that it is considerably more data-efficient to train semantic state-space models due to the lack of need to teach the model to pay attention to the suitable patches of earlier context.
Ultimately, we see no reason why SSSMs could not match the performance of full-history-accumulating models, which begs the question: Why train full-history-accumulating models in the first place?

\section{Conclusion}
\label{section:conclusion}

Our experiments provide affirmative answers to all three research questions \cref{rq:long_horizon,rq:long_input,rq:single_action}.
In particular, we found that on long-horizon tasks, even simple end-of-conversation compaction usually matches or slightly improves on full-history accuracy, with only modest losses in the remaining settings.
We showed that on long-input tasks, selective input reading with compaction as a fallback matches or improves (sometimes considerably) the full-input baselines in almost every case while keeping live context short.
Finally, single-action compaction nearly preserves or improves on the baseline accuracy while carrying only a compact semantic state between actions.

\cref{rq:long_horizon,rq:long_input,rq:single_action} were motivated by the root question \Cref{rq:core_question}: ``Are long contexts \textit{necessary}?''
Based on the affirmative answers to all of the above, we conclude that long contexts are not necessary for the completion of ``long-context'' tasks.
This is far from surprising.
After all, the very motivation behind the definition of computability rests on the assumption that anything computable may be computed by comprehending only a bounded amount of input at a time and maintaining only a finite state of mind \cite{turing}.

Our evidence does not rule out the existence of tasks for which long context is beneficial, nor does it prove that every possible task can be solved with short context.
It does, however, show that the standard arguments in favor of long contexts stemming from long input and long conversational horizons are insufficient: disciplined state maintenance and selective access remove the presumed necessity in every setting we tested.
We believe that with the appropriate adjustments made to the post-training of LMs, the context window size requirements currently imposed on these models by the standard arguments for long contexts could be removed, and model context window sizes could shrink drastically, significantly improving the cost of token generation on the cloud and the speed of token generation on local inference accelerators.

\bibliographystyle{plainnat}
\bibliography{bibliography}

\clearpage
\appendix
\crefalias{section}{appendix}
\section{What characterizes a good compaction?}
\label{appendix:good_compaction}

It is a common experience of human-in-the-loop agent-aided SWE workflows that after a compaction is performed, certain crucial information disappears from the context and has to be obtained again, slowing down the SWE process or perhaps even confusing the model and losing crucial user preferences included in previous instructions.
Note that human-in-the-loop agent-aided SWE workflows consist of two parts: autonomous SWE work performed by the agent, and reception and comprehension of the human's steering feedback.
Because of this, compactions produced for such workflows need to account for not only the information the agent needs to achieve its goals, but also for all potential information that might be relevant to follow-up autonomous work that might be triggered as a result of the human's feedback.

In our experimentation, we initially set out to identify the cheapest, most useful method of compaction that can preserve as much task accuracy as possible on autonomous SWE work.
We discovered that treating compaction as a mere special case of summarization is suboptimal for this purpose.
LM summarizations generally attempt to give equal weight to all facts present in the input and are prone to overemphasize repeated segments of the agent trace regardless of their ultimate relevance to the tail of the trace being compacted.
We explored the subject along two axes: the method (e.g., summarization or message pruning) and the information being pruned (e.g., the entire trace or only the trace segment since the last compaction plus the contents of the last compaction).
We discovered that the best compactions are not those that faithfully capture the gist of the agentic trace, but those that best select facts relevant to the work yet to be performed -- i.e., they are not narratives of the execution history but carefully cherry-picked briefs for further execution. 
In particular, we found that the most accuracy-preserving compactions, regardless of the frequency of compaction, are those that contain (1) a verbatim or near-verbatim copy of the input task, (2) a record of agent's current occupation, intentions, and planned next steps, and (3) all the artifacts that are of direct relevance to future execution, including (3a) a decomposed plan made from the input task, (3b) a list of requirements detailing when to consider the task complete, and any pieces of information or pointers to artifacts containing information that is of relevance to the aforementioned two.
Finally, we found that the above structure of compactions readily extends even to workflows that are not necessarily SWE-related in their nature.
We used the same compaction structure for every type of workflow considered in this work.

\section{Estimated optimized retail cost}
\label{appendix:retail_cost_fit}

For model $M$, let $C_{\text{output},M}(W)$ denote the USD per million output tokens at size $W$.
For measured models, let $T_M(W)$ be the peak aggregate decode throughput (output tokens per second after input-context processing, assuming KV cached input). Fixed GPU allocation and hourly rate imply the $16$K-to-$256$K cost ratio
\begin{equation}
  r_M=\frac{C_{\text{output},M}(16\mathrm{K})}{C_{\text{output},M}(256\mathrm{K})}
     =\frac{T_M(256\mathrm{K})}{T_M(16\mathrm{K})}
     =\begin{cases}
       0.2772 & \text{GLM 5.2},\\
       0.1777 & \text{Kimi K2.6},\\
       0.1439 & \text{Qwen 3 Coder Next}.
     \end{cases}
  \label{equation:retail_cost_ratio}
\end{equation}

We take the rounded mean of the above measurements, $r=0.20$.

For target frontier model $M$, let $W^{\text{max}}_{M}$ be the assumed reference context window size: $200,000$ tokens for Opus~4.5 and $272$K for GPT 5.6 Sol~\cite{anthropic2026opus45pricing,openai2026gpt56pricing}.
The GPT's $272$K reference context window size is its standard-price ceiling (the model supports $1,050,000$ tokens, but our compaction $W$ never exceeds the standard-price ceiling).
For each compaction experiment, let $W\leq W^{\text{max}}_{M}$ be the EoCC total context limit and let $W^{\text{max}}_{M}$ be the context window size for the FHA baseline (\Cref{section:long_conversations}).
To estimate the cost of output $C^{\text{estimate}}_{\text{output},M}(W)$ of model $M$, we anchor $C_{\text{output},M}(W^{\text{max}}_{M})$ at the standard retail output price, \$25 for Opus~4.5 or \$20 for GPT 5.6 Sol, and then interpolate to $r$-times-the-standard-price at $16$K, clamping smaller $W$s:
\begin{equation}
  C^{\text{est.}}_{\text{output},M}(W)=C_{\text{output},M}(W^{\text{max}}_{M})
  \left[r+(1-r)\frac{\max(W,16\mathrm{K})-16\mathrm{K}}{W^{\text{max}}_{M}-16\mathrm{K}}\right].
  \label{equation:retail_cost_context_factor}
\end{equation}

Let $C_{\text{input},M}(W)$ and $C_{\text{cached},M}(W)$ be the est. USD per million uncached-input and cached-input tokens, respectively.
These we deduce by preserving the original 5:1 output-to-uncached-input ratio and 10:1 uncached-input-to-cached-input ratio of the frontier models:
\begin{equation}
  C_{\text{input},M}(W)=\frac{C^{\text{est.}}_{\text{output},M}(W)}{5},\qquad
  C_{\text{cached},M}(W)=\frac{C^{\text{est.}}_{\text{input},M}(W)}{10}.
  \label{equation:retail_cost_input_ratios}
\end{equation}

For every experiment with model $M$ maintaining the EoCC total context limit at $W$, let $I$, $H$, and $O$ be aggregate millions of uncached input, cached input (cache hits), and generated output tokens, respectively.
We compute the estimate of the final \emph{estimated optimized retail cost} as reported in the paper as:
\begin{equation}
  {EORC}_M(W)=I C^{\text{est.}}_{\text{input},M}(W)
    +H C^{\text{est.}}_{\text{cached},M}(W)+O C^{\text{est.}}_{\text{output},M}(W).
  \label{equation:retail_cost_estimate}
\end{equation}

\clearpage
\section{Detailed long-horizon results}
\label{appendix:long_horizon_results}

The tables below report accuracy and cost profiles for Claude Opus~4.5 and GPT 5.6 Sol.
Each table compares a baseline FHA agent (\Cref{section:harnesses}) against an EoCC agent with different total context limits and compaction thresholds (\Cref{section:long_conversations}).
Values are written as ``total context limit / compaction threshold'' in tokens: the agent compacts when the running context approaches the compaction threshold and never exceeds the total context limit.

Throughout the results, notice the following:
\begin{itemize}
  \item In terms of accuracy, compactions tend to perform closely on-par with or slightly better than the corresponding baseline.
  \item In terms of total token spend, new input token counts generally increase with decreasing context window size while the re-used cached tokens generally decrease, consistently with the intuitive expectation. Output tokens are generally elevated when compacting.
\end{itemize}

Inspecting \Cref{table:swe_profile}, we observe that every evaluated pair of total context limit and compaction threshold matches or improves on the baseline for GPT 5.6 Sol; Claude Opus~4.5 improves at every setting except the tightest ($16$K / $12$K), where accuracy decreases by 1.2 percentage points.
The best results improve accuracy by 4.8 percentage points for Claude Opus~4.5 and 4.4 points for GPT 5.6 Sol.

\begin{table}[H]
\centering
\captiontable{%
\resizebox{\textwidth}{!}{%
\begin{tabular}{l cc cc cc cc cc cc}
\toprule
 & \multicolumn{2}{c}{Accuracy} & \multicolumn{2}{c}{Token spend (M)} & \multicolumn{2}{c}{\shortstack{Nominal retail\\cost (USD)}} & \multicolumn{2}{c}{\shortstack{Estimated optimized\\retail cost (USD)}} & \multicolumn{2}{c}{Compactions} & \multicolumn{2}{c}{Total length (K)} \\
\cmidrule(lr){2-3}\cmidrule(lr){4-5}\cmidrule(lr){6-7}\cmidrule(lr){8-9}\cmidrule(lr){10-11}\cmidrule(lr){12-13}
Configuration & Opus & GPT & Opus & GPT & Opus & GPT & Opus & GPT & Opus & GPT & Opus & GPT \\
\midrule
Baseline (no compaction) & 70.0 & 84.4 & 20.0\,/\,420.0\,/\,4.4 & 16.4\,/\,100.2\,/\,3.1 & \$420.00 & \$167.68 & \$420.00 & \$167.68 & 0\,/\,0\,/\,0 & 0\,/\,0\,/\,0 & 48\,/\,32\,/\,200 & 22\,/\,18\,/\,173 \\
Compaction, $48$K / $36$K & 73.6 & 88.8 & 20.0\,/\,334.7\,/\,4.5 & 17.6\,/\,94.2\,/\,3.9 & \$379.85 & \$186.08 & \$130.20 & \$55.82 & 0.3\,/\,0\,/\,6 & 0.4\,/\,0\,/\,5 & 30\,/\,25\,/\,226 & 26\,/\,22\,/\,156 \\
Compaction, $32$K / $24$K & 72.0 & 85.0 & 23.0\,/\,255.9\,/\,4.7 & 17.2\,/\,73.4\,/\,4.3 & \$360.45 & \$184.16 & \$97.82 & \$46.04 & 0.8\,/\,1\,/\,13 & 0.9\,/\,1\,/\,9 & 32\,/\,28\,/\,336 & 28\,/\,24\,/\,189 \\
Compaction, $24$K / $18$K & 74.8 & 85.0 & 22.0\,/\,166.5\,/\,4.1 & 17.7\,/\,60.4\,/\,4.7 & \$295.75 & \$188.96 & \$69.71 & \$42.52 & 1.0\,/\,1\,/\,18 & 1.5\,/\,1\,/\,18 & 29\,/\,23\,/\,362 & 32\,/\,24\,/\,287 \\
Compaction, $16$K / $12$K & 68.8 & 84.6 & 28.3\,/\,127.3\,/\,4.3 & 19.7\,/\,45.8\,/\,5.3 & \$312.65 & \$203.12 & \$62.53 & \$40.62 & 2.1\,/\,1\,/\,32 & 3.0\,/\,2\,/\,61 & 34\,/\,24\,/\,415 & 36\,/\,26\,/\,605 \\
\bottomrule
\end{tabular}}}{
  \textbf{SWE-bench Verified.}
  \emph{Accuracy} is the percentage of instances \emph{resolved}.
  \emph{Token spend} lists new\,/\,cached input and output, expressed in M tokens, aggregated
  over the benchmark (\emph{new} excludes prompt-cache hits).
  \emph{Nominal retail cost} multiplies these rounded totals by published standard API prices per million new input\,/\,cached input\,/\,output tokens: \$5\,/\,\$0.50\,/\,\$25 for Opus~4.5 and \$4\,/\,\$0.40\,/\,\$20 for GPT 5.6 Sol, as of September 29, 2026~\cite{anthropic2026opus45pricing,openai2026gpt56pricing}.
  \emph{Estimated optimized retail cost} applies the context factor defined in \Cref{appendix:retail_cost_fit} to the same token totals.
  \emph{Compactions} and
  \emph{Total length} are per task, given as mean\,/\,median\,/\,max; \emph{Total length} is in K tokens and sums each sub-conversation's final context across
  stretches.
  \label{table:swe_profile}
}
\end{table}

On SWE-bench Multilingual (\Cref{table:swe_multilingual_profile}), the largest accuracy gains occur at $48$K / $36$K: 1.7 percentage points for Claude Opus~4.5 and 0.7 points for GPT 5.6 Sol.
At the tightest setting ($16$K / $12$K), accuracy falls below the corresponding baselines by 2.0 and 1.6 percentage points, respectively.

\begin{table}[H]
\centering
\captiontable{%
\resizebox{\textwidth}{!}{%
\begin{tabular}{l cc cc cc cc cc cc}
\toprule
 & \multicolumn{2}{c}{Accuracy} & \multicolumn{2}{c}{Token spend (M)} & \multicolumn{2}{c}{\shortstack{Nominal retail\\cost (USD)}} & \multicolumn{2}{c}{\shortstack{Estimated optimized\\retail cost (USD)}} & \multicolumn{2}{c}{Compactions} & \multicolumn{2}{c}{Total length (K)} \\
\cmidrule(lr){2-3}\cmidrule(lr){4-5}\cmidrule(lr){6-7}\cmidrule(lr){8-9}\cmidrule(lr){10-11}\cmidrule(lr){12-13}
Configuration & Opus & GPT & Opus & GPT & Opus & GPT & Opus & GPT & Opus & GPT & Opus & GPT \\
\midrule
Baseline (no compaction) & 66.3 & 79.3 & 11.9\,/\,384.1\,/\,3.0 & 7.4\,/\,66.2\,/\,1.6 & \$326.55 & \$88.08 & \$326.55 & \$88.08 & 0\,/\,0\,/\,0 & 0\,/\,0\,/\,0 & 35\,/\,31\,/\,95 & 23\,/\,21\,/\,63 \\
Compaction, $48$K / $36$K & 68.0 & 80.0 & 12.8\,/\,245.5\,/\,2.9 & 8.8\,/\,65.6\,/\,1.9 & \$259.25 & \$99.44 & \$87.56 & \$29.48 & 0.4\,/\,0\,/\,3 & 0.3\,/\,0\,/\,5 & 38\,/\,33\,/\,157 & 28\,/\,23\,/\,190 \\
Compaction, $32$K / $24$K & 67.0 & 78.7 & 14.0\,/\,147.0\,/\,2.8 & 10.0\,/\,49.1\,/\,2.3 & \$213.50 & \$105.64 & \$57.23 & \$26.16 & 1.1\,/\,1\,/\,8 & 1.1\,/\,1\,/\,14 & 39\,/\,32\,/\,192 & 33\,/\,27\,/\,312 \\
Compaction, $24$K / $18$K & 66.0 & 78.7 & 15.6\,/\,91.0\,/\,2.7 & 10.8\,/\,40.3\,/\,2.6 & \$191.00 & \$111.32 & \$44.54 & \$24.85 & 1.9\,/\,1\,/\,13 & 1.8\,/\,1\,/\,18 & 41\,/\,31\,/\,235 & 37\,/\,28\,/\,312 \\
Compaction, $16$K / $12$K & 64.3 & 77.7 & 17.0\,/\,65.4\,/\,2.7 & 12.4\,/\,31.5\,/\,3.1 & \$185.20 & \$124.20 & \$37.04 & \$24.84 & 2.7\,/\,2\,/\,26 & 3.4\,/\,2\,/\,44 & 42\,/\,33\,/\,340 & 43\,/\,30\,/\,472 \\
\bottomrule
\end{tabular}}}{
  \textbf{SWE-bench Multilingual.} Results for Claude Opus~4.5 High and GPT~5.6 Sol Medium on 300 instances per model.
  Columns and quantities are as in \Cref{table:swe_profile}.\label{table:swe_multilingual_profile}}
\end{table}

The Terminal-Bench results in \Cref{table:terminal_profile} show a similar pattern: compaction improves Claude Opus~4.5 at every evaluated pair of total context limit and compaction threshold. GPT 5.6 Sol improves by 1.2 percentage points at $48$K / $36$K and loses 1.1, 2.2, and 2.2 points at $32$K / $24$K, $24$K / $18$K, and $16$K / $12$K, respectively.
The largest gain is 5.6 percentage points for Claude Opus~4.5 with a $32$K total context limit and $24$K compaction threshold.

\begin{table}[H]
\centering
\captiontable{%
\resizebox{\textwidth}{!}{%
\begin{tabular}{l cc cc cc cc cc cc}
\toprule
 & \multicolumn{2}{c}{Accuracy} & \multicolumn{2}{c}{Token spend (M)} & \multicolumn{2}{c}{\shortstack{Nominal retail\\cost (USD)}} & \multicolumn{2}{c}{\shortstack{Estimated optimized\\retail cost (USD)}} & \multicolumn{2}{c}{Compactions} & \multicolumn{2}{c}{Total length (K)} \\
\cmidrule(lr){2-3}\cmidrule(lr){4-5}\cmidrule(lr){6-7}\cmidrule(lr){8-9}\cmidrule(lr){10-11}\cmidrule(lr){12-13}
Configuration & Opus & GPT & Opus & GPT & Opus & GPT & Opus & GPT & Opus & GPT & Opus & GPT \\
\midrule
Baseline (no compaction) & 47.2 & 87.6 & 3.9\,/\,81.6\,/\,1.1 & 3.6\,/\,30.4\,/\,1.1 & \$87.80 & \$48.56 & \$87.80 & \$48.56 & 0\,/\,0\,/\,0 & 0\,/\,0\,/\,0 & 34\,/\,27\,/\,130 & 28\,/\,17\,/\,133 \\
Compaction, $48$K / $36$K & 50.6 & 88.8 & 6.2\,/\,77.5\,/\,1.8 & 3.7\,/\,16.7\,/\,1.3 & \$114.75 & \$47.48 & \$39.33 & \$14.24 & 0.9\,/\,0\,/\,7 & 0.6\,/\,0\,/\,8 & 51\,/\,29\,/\,288 & 33\,/\,18\,/\,281 \\
Compaction, $32$K / $24$K & 52.8 & 86.5 & 6.7\,/\,47.1\,/\,1.7 & 4.0\,/\,14.1\,/\,1.6 & \$99.55 & \$53.64 & \$27.02 & \$13.41 & 1.7\,/\,1\,/\,13 & 1.4\,/\,0\,/\,18 & 51\,/\,31\,/\,296 & 37\,/\,18\,/\,359 \\
Compaction, $24$K / $18$K & 50.6 & 85.4 & 6.5\,/\,34.0\,/\,1.7 & 6.8\,/\,18.0\,/\,2.8 & \$92.00 & \$90.40 & \$21.68 & \$20.34 & 2.1\,/\,1\,/\,14 & 4.0\,/\,2\,/\,20 & 48\,/\,25\,/\,271 & 72\,/\,44\,/\,331 \\
Compaction, $16$K / $12$K & 48.3 & 85.4 & 7.6\,/\,23.0\,/\,1.6 & 4.2\,/\,10.3\,/\,2.0 & \$89.50 & \$60.92 & \$17.90 & \$12.18 & 3.5\,/\,1\,/\,25 & 3.8\,/\,2\,/\,29 & 49\,/\,23\,/\,312 & 43\,/\,23\,/\,297 \\
\bottomrule
\end{tabular}}}{\textbf{Terminal-Bench (TB-2.1).} Columns and quantities are as in \Cref{table:swe_profile}; \emph{Accuracy} is the percentage of tasks solved.\label{table:terminal_profile}}
\end{table}

As shown in \Cref{table:frames_profile}, on FRAMES, compaction reduces accuracy by 1.3--3.7 percentage points for Claude Opus~4.5 and 1.5--3.3 points for GPT 5.6 Sol relative to their updated baselines of 81.0\% and 90.2\%.
At $16$K / $12$K, the reductions are 1.4 and 3.1 points, while estimated optimized costs fall by $7.7\times$ and $3.8\times$, respectively.

\begin{table}[H]
\centering
\captiontable{%
\resizebox{\textwidth}{!}{%
\begin{tabular}{l cc cc cc cc cc cc}
\toprule
 & \multicolumn{2}{c}{Accuracy} & \multicolumn{2}{c}{Token spend (M)} & \multicolumn{2}{c}{\shortstack{Nominal retail\\cost (USD)}} & \multicolumn{2}{c}{\shortstack{Estimated optimized\\retail cost (USD)}} & \multicolumn{2}{c}{Compactions} & \multicolumn{2}{c}{Total length (K)} \\
\cmidrule(lr){2-3}\cmidrule(lr){4-5}\cmidrule(lr){6-7}\cmidrule(lr){8-9}\cmidrule(lr){10-11}\cmidrule(lr){12-13}
Configuration & Opus & GPT & Opus & GPT & Opus & GPT & Opus & GPT & Opus & GPT & Opus & GPT \\
\midrule
Baseline (no compaction) & 81.0 & 90.2 & 27.4\,/\,606.5\,/\,2.7 & 21.9\,/\,177.6\,/\,2.4 & \$507.75 & \$206.64 & \$507.75 & \$206.64 & 0\,/\,0\,/\,0 & 0\,/\,0\,/\,0 & 25\,/\,14\,/\,92 & 17\,/\,10\,/\,140 \\
Compaction, $48$K / $36$K & 79.7 & 88.7 & 30.2\,/\,307.6\,/\,2.6 & 27.6\,/\,171.1\,/\,3.5 & \$369.80 & \$248.84 & \$126.76 & \$74.65 & 0.4\,/\,0\,/\,3 & 0.5\,/\,0\,/\,7 & 24\,/\,12\,/\,103 & 26\,/\,10\,/\,206 \\
Compaction, $32$K / $24$K & 79.4 & 88.0 & 28.7\,/\,196.6\,/\,2.4 & 30.0\,/\,147.7\,/\,4.2 & \$301.80 & \$263.08 & \$81.90 & \$65.77 & 0.8\,/\,0\,/\,4 & 1.1\,/\,0\,/\,9 & 27\,/\,13\,/\,111 & 30\,/\,10\,/\,177 \\
Compaction, $24$K / $18$K & 77.3 & 86.9 & 36.8\,/\,173.2\,/\,2.9 & 29.4\,/\,125.4\,/\,4.6 & \$343.10 & \$259.76 & \$80.87 & \$58.45 & 1.0\,/\,0\,/\,6 & 1.6\,/\,0\,/\,11 & 26\,/\,12\,/\,112 & 30\,/\,9\,/\,164 \\
Compaction, $16$K / $12$K & 79.6 & 87.1 & 39.2\,/\,108.2\,/\,3.1 & 28.9\,/\,105.2\,/\,5.7 & \$327.60 & \$271.68 & \$65.52 & \$54.34 & 1.6\,/\,1\,/\,9 & 2.6\,/\,1\,/\,13 & 25\,/\,13\,/\,104 & 31\,/\,11\,/\,137 \\
\bottomrule
\end{tabular}}}{\textbf{FRAMES.} Columns and quantities are as in \Cref{table:swe_profile}; \emph{Accuracy} is the percentage of questions answered correctly.\label{table:frames_profile}}
\end{table}

On the 200 long-context multi-turn episodes of BFCLv3, \Cref{table:bfcl_profile}
reports accuracy, token consumption, compaction frequency and accumulated total context.

\begin{table}[H]
\centering
\captiontable{%
\resizebox{\textwidth}{!}{%
\begin{tabular}{l cc cc cc cc cc cc}
\toprule
 & \multicolumn{2}{c}{Accuracy} & \multicolumn{2}{c}{Token spend (M)} & \multicolumn{2}{c}{\shortstack{Nominal retail\\cost (USD)}} & \multicolumn{2}{c}{\shortstack{Estimated optimized\\retail cost (USD)}} & \multicolumn{2}{c}{Compactions} & \multicolumn{2}{c}{Total length (K)} \\
\cmidrule(lr){2-3}\cmidrule(lr){4-5}\cmidrule(lr){6-7}\cmidrule(lr){8-9}\cmidrule(lr){10-11}\cmidrule(lr){12-13}
Configuration & Opus & GPT & Opus & GPT & Opus & GPT & Opus & GPT & Opus & GPT & Opus & GPT \\
\midrule
Baseline (no compaction) & 62.0 & 39.5 & 3.6\,/\,18.9\,/\,0.3 & 1.8\,/\,15.0\,/\,0.1 & \$34.95 & \$15.20 & \$34.95 & \$15.20 & 0.0\,/\,0\,/\,0 & 0.0\,/\,0\,/\,0 & 15.2\,/\,12.5\,/\,70.9 & 11.2\,/\,7.2\,/\,66.5 \\
Compaction, $24$K / $18$K & 64.0 & 39.0 & 4.6\,/\,12.6\,/\,0.3 & 2.2\,/\,9.5\,/\,0.2 & \$36.80 & \$16.60 & \$8.64 & \$3.74 & 0.3\,/\,0\,/\,3 & 0.2\,/\,0\,/\,2 & 17.3\,/\,11.5\,/\,90.1 & 12.3\,/\,7.2\,/\,77.7 \\
Compaction, $16$K / $12$K & 61.5 & 39.0 & 4.5\,/\,8.9\,/\,0.3 & 2.0\,/\,8.1\,/\,0.2 & \$34.45 & \$15.24 & \$6.89 & \$3.05 & 0.6\,/\,1\,/\,4 & 0.4\,/\,0\,/\,3 & 19.7\,/\,18.8\,/\,104.1 & 14.5\,/\,7.4\,/\,77.8 \\
Compaction, $12$K / $8$K & 62.5 & 39.5 & 4.9\,/\,7.5\,/\,0.4 & 2.2\,/\,6.9\,/\,0.3 & \$38.25 & \$17.56 & \$7.65 & \$3.51 & 1.1\,/\,1\,/\,5 & 0.6\,/\,1\,/\,4 & 22.2\,/\,18.6\,/\,89.6 & 15.2\,/\,11.5\,/\,84.0 \\
\bottomrule
\end{tabular}}}{\textbf{BFCLv3 (200 long-context multi-turn episodes).} Columns and quantities are as in \Cref{table:swe_profile}; \emph{Accuracy} is the percentage of episodes passed.
\label{table:bfcl_profile}}
\end{table}

\Needspace{16\baselineskip}
Finally, \Cref{table:tau2_profile} shows that compaction keeps Claude Opus~4.5 within 1.1 percentage points of its 79.3\% baseline accuracy on $\tau^2$-bench: accuracy is 78.2\% with a $12$K total context limit and $8$K compaction threshold, matches the baseline with $16$K / $12$K, and reaches 78.8\% with $8$K / $6$K. GPT 5.6 Sol improves with a $12$K total context limit and $8$K compaction threshold and remains near its baseline with the other total context limits and compaction thresholds.

\begin{table}[H]
\centering
\captiontable{%
\resizebox{\textwidth}{!}{%
\begin{tabular}{l cc cc cc cc}
\toprule
 & \multicolumn{2}{c}{Accuracy} & \multicolumn{2}{c}{Token spend (M)} & \multicolumn{2}{c}{\shortstack{Nominal retail\\cost (USD)}} & \multicolumn{2}{c}{\shortstack{Estimated optimized\\retail cost (USD)}} \\
\cmidrule(lr){2-3}\cmidrule(lr){4-5}\cmidrule(lr){6-7}\cmidrule(lr){8-9}
Configuration & Opus & GPT & Opus & GPT & Opus & GPT & Opus & GPT \\
\midrule
Baseline (no compaction) & 79.3 & 73.9 & 14.5\,/\,0.3 & 11.4\,/\,0.2 & \$80.00 & \$49.60 & \$80.00 & \$49.60 \\
Compaction, $16$K / $12$K & 79.3 & 74.5 & 15.1\,/\,0.3 & 15.3\,/\,0.2 & \$83.00 & \$65.20 & \$16.60 & \$13.04 \\
Compaction, $12$K / $8$K & 78.2 & 77.2 & 14.3\,/\,0.3 & 15.2\,/\,0.3 & \$79.00 & \$66.80 & \$15.80 & \$13.36 \\
Compaction, $8$K / $6$K & 78.8 & 73.9 & 13.2\,/\,0.3 & 14.4\,/\,0.3 & \$73.50 & \$63.60 & \$14.70 & \$12.72 \\
\bottomrule
\end{tabular}}}{\textbf{$\tau^2$-bench.} Columns and quantities are as in \Cref{table:swe_profile}.\label{table:tau2_profile}}
\end{table}

\clearpage
\section{Open-weight model results}
\label{appendix:open_weight_results}

We also evaluate the five quantized open-weight models described in \Cref{section:models} on SWE-bench Verified and Terminal-Bench 2.1 (\Cref{table:open_weight_swe,table:open_weight_terminal}).
The compaction settings are $48$K / $40$K, $32$K / $24$K, and $16$K / $12$K, expressed as total context limit / compaction threshold.

Robustness to compaction varies across models.
Qwen 3.6 27B and MiniMax M2.7 are comparatively robust, whereas Gemma 4 31B, Step 3.7 Flash, and especially Qwen 3 Coder Next show larger losses as the context budget tightens.
Qwen 3 Coder Next catastrophically collapses on SWE-bench Verified from 53.0\% to 3.6\% at $16$K / $12$K, mostly because its verbose trajectories leave too little room for useful progress between compactions.
At this tightest setting, every model loses SWE-bench accuracy relative to its baseline, although Qwen 3.6 27B and MiniMax M2.7 retain their Terminal-Bench baseline scores.

Observe that despite the gradual tightening of the compaction threshold, all models maintain their ability to complete a significant fraction of the benchmark workflows successfully.
This goes on to show that even small- to medium-sized compact coding models with no explicit compaction post-training exhibit material utility while operating with shorter context window sizes.

\begin{table}[H]
\centering
\captiontable{%
\begin{tabular}{lrrrr}
\toprule
Model & Baseline & $48$K / $40$K & $32$K / $24$K & $16$K / $12$K \\
\midrule
Qwen 3.6 27B & 68.6 & 69.4 & 68.2 & 65.0 \\
Qwen 3 Coder Next & 53.0 & 49.2 & 40.4 & 3.6 \\
Gemma 4 31B & 65.2 & 65.0 & 65.8 & 58.0 \\
Step 3.7 Flash & 70.0 & 66.8 & 64.6 & 56.6 \\
MiniMax M2.7 & 72.6 & 73.0 & 73.2 & 67.8 \\
\bottomrule
\end{tabular}}{\textbf{Open-weight results on SWE-bench Verified.} Scores are percentages of 500 instances resolved; patch-application failures count as unresolved. \emph{Baseline} uses full-history accumulation without an agent context cap; the other columns report EoCC at the indicated total context limit / compaction threshold.\label{table:open_weight_swe}}
\end{table}

\begin{table}[H]
\centering
\captiontable{%
\begin{tabular}{lrrrr}
\toprule
Model & Baseline & $48$K / $40$K & $32$K / $24$K & $16$K / $12$K \\
\midrule
Qwen 3.6 27B & 43.8 & 48.3 & 49.4 & 43.8 \\
Qwen 3 Coder Next & 29.2 & 34.8 & 34.8 & 30.3 \\
Gemma 4 31B & 40.4 & 34.8 & 29.2 & 33.7 \\
Step 3.7 Flash & 41.6 & 32.6 & 34.8 & 28.1 \\
MiniMax M2.7 & 40.4 & 37.1 & 38.2 & 40.4 \\
\bottomrule
\end{tabular}}{\textbf{Open-weight results on Terminal-Bench 2.1.} Columns are as in \Cref{table:open_weight_swe}; scores are percentages of 89 tasks solved.\label{table:open_weight_terminal}}
\end{table}

\clearpage
\section{Selective reading of long inputs}
\label{appendix:selective_reading}
\begingroup
\lstset{
  basicstyle=\ttfamily\footnotesize,
  breaklines=true,
  columns=fullflexible,
  keepspaces=true,
  showstringspaces=false,
  frame=none,
  postbreak={},
  aboveskip=6pt,
  belowskip=6pt
}

Our generalized adaptation of \texttt{mini-swe-agent}, \texttt{mini-agent}, exposes long benchmark inputs as text files that the agent can inspect through a bash tool.
LongBench~v2 uses one document file, whereas LOONG and AA-LCR expose multiple source files.
The operation of \texttt{mini-agent} is straightforward.
The model receives the question and file information and then chooses which text to retrieve in order to start answering the question.
A succession of Bash-invoked searches then scan entire files, but only their returned matched excerpts enter the model's context.

The basic process of a single search is to identify the entities, quantities, or claims needed for the answer, and then to read a small surrounding window to interpret the candidate evidence.
Further related searches might then be used to retrieve more candidates looked up using broader sets of synonyms.
Standard commands such as \texttt{grep}, \texttt{sed}, \texttt{head}, and \texttt{tail} are the backbone of the search process.
Convenience commands provide document sizes (\texttt{doc-stats}), file manifests (\texttt{docs-list}), bounded reads (\texttt{doc-read}), and search-term variants (\texttt{doc-syn}) allow for more surgical readouts from files and expanded search follow-ups.

\subsection{Instructions supplied to the agent.}
The base system prompt gives the following \texttt{man}-like tool description:
\begin{lstlisting}
doc-read [--start CHAR] [--chars N] [--lines A:B]
    Read a BOUNDED window (default 6000 chars, hard-capped at 8000). Give a
    character offset window, or a 1-based inclusive line range. The header
    tells you the command to read the next contiguous window. This is how you
    read the document sequentially without ever exceeding your budget.

grep -n -i -E '<regex>' "$LONGBENCH_DOC"
    Find where a name / number / phrase appears (with -n for line numbers).
    Then doc-read around those lines. This is usually far faster than reading
    straight through -- locate the relevant passages, then read only those.

...
\end{lstlisting}
\noindent

The system prompt also makes the preference for selective reading explicit:
\begin{lstlisting}
IMPORTANT --- read only what you need.
Keep each read small (<= ~6000 chars).
Do NOT try to dump the whole document.
Search for the entities/claims each option hinges on, read the surrounding passages, and verify.
\end{lstlisting}
\noindent

\subsection{Examples from recorded runs.}
The following excerpts retain only commands and selected tool output.
They are illustrative traces, not estimates of typical reading cost.

\medskip\noindent\textbf{1. LongBench~v2: locate a word, then read its grammatical context.}
In the question of the dataset's first sample, the agent translates a Kalamang sentence using a grammar of roughly $1$M characters ($19$K lines).
It searches for a word from the sentence and uses the returned line number to select a 36-line passage.
\begin{lstlisting}
$ grep -n -i -E '\bkademor\b' "$LONGBENCH_DOC" | head -20
3840:Two other verbs ewa 'to speak' and kademor 'to be mad' introduce an object to the scene when the applicative is used. [...]
[...]

$ doc-read --lines 3835:3870 "$LONGBENCH_DOC"
[lines 3835:3870 of 18830]
[...]
Two other verbs ewa 'to speak' and kademor 'to be mad' introduce an object to the scene when the applicative is used. [...]
mena ma ecie ma an=at ko=ewa
otherwise 3sg return 3sg 1sg=obj appl=speak
'Otherwise when he comes back, he is mad at me.' *-0000-0000-0004-1BA3-3[conv100:15]
[...]
\end{lstlisting}
Through a follow-up \texttt{doc-read}, the agent retrieves surrounding grammar and examples, rather than relying solely on the lexical match retrieved with the earlier \texttt{grep}.

\medskip\noindent\textbf{2. LOONG: search across papers, then verify a paper's identity.}
The question of the dataset's first sample asks for citation relationships among three supplied papers (roughly $138$K characters in total).
The agent searches two other papers for the target paper's name.
Later, it reads only the beginning of the matching paper to verify its title.
\begin{lstlisting}
$ grep -rn -i -E "OpenMoE" doc02.txt doc03.txt
doc02.txt:279:Fuzhao Xue, Zian Zheng, Yao Fu, Jinjie Ni, Zangwei Zheng, Wangchunshu Zhou, and Yang You. OpenMoE: An early effort on open mixture-of-experts language models. arXiv preprint arXiv:2402.01739, 2024.

[... intervening checks omitted ...]

$ doc-read doc02.txt --start 0 --chars 1000
[doc02.txt chars 0:1000 of 44964]  next: doc-read doc02.txt --start 1000 --chars 1000
Lory: Fully Differentiable Mixture-of-Experts for Autoregressive Language Model Pre-training
[...]
\end{lstlisting}
This example combines a \texttt{grep} search across files with a surgical $1$K-character follow-up read.

\medskip\noindent\textbf{3. AA-LCR: read the context needed to apply an exclusion.}
The question of the second sample asks for industries ranked by consumer-related infringements over three decades, explicitly excluding broadcasting.
Four supplied documents contain roughly $404$K characters.
After narrowing the search to one report, the \texttt{mini-agent} retrieves a short passage containing the industry names, counts, and time span.
\begin{lstlisting}
$ grep -n -i -E 'consumer infringements|consumer-related|infringement notices|enforceable undertakings|Broadcasting' documents/doc02.txt
259:There have been 12 infringement notices and enforceable undertakings issued
267:example, in the Broadcasting industry, a single firm received 10 contraventions
[...]

[... intervening checks omitted ...]

$ doc-read documents/doc02.txt --lines 245:272; grep -n -i -E 'There have been [0-9]+|only [0-9]+ infringements|received [0-9]+ contraventions' documents/doc02.txt
[documents/doc02.txt lines 245:272 of 657]
[...]
There have been 12 infringement notices and enforceable undertakings issued
over 30 years in the airline industry, which is famously dominated by a small
number of companies.

By contrast, in the Accommodation industry, which is less concentrated and has
a higher number businesses, only 4 infringements were issued.

Repeat offences in concentrated industries are a key driver of these results. For
example, in the Broadcasting industry, a single firm received 10 contraventions
over the 30 year period (see Figure A11 for more information).

[...]
\end{lstlisting}
Here, a sequence of \texttt{grep} searches and \texttt{doc-read}s is executed until the key relevant passage has been identified.

\endgroup

\clearpage
\section{Detailed long-input results}
\label{appendix:long_input_results}

To demonstrate \Cref{rq:long_input}, we compare the performance of selectively-reading, compacting agents against the more straightforward benchmark baselines that involve pasting long inputs into the models' work conversations.
Following the long-input accommodations in \Cref{section:long_inputs}, the agents tested read only task-relevant portions of each document and use compaction when their working context reaches the compaction threshold.
Compaction thresholds are reported as ``total context limit / compaction threshold'' in tokens.

LongBench v2 shows that selective input reading preserves baseline-level accuracy without placing the full documents in the models' working context.
In \Cref{table:longbench_profile}, GPT 5.6 Sol improves by 1.8 percentage points and Claude Opus~4.5 remains within 0.4 points of its baseline.
The $16$K total context limit bounds peak live context, while compaction events remain infrequent.

\begin{table}[H]
\centering
\captiontable{%
\resizebox{\textwidth}{!}{%
\begin{tabular}{l cc cc cc cc cc}
\toprule
 & \multicolumn{2}{c}{Accuracy} & \multicolumn{2}{c}{Token spend (M)} & \multicolumn{2}{c}{\shortstack{Nominal retail\\cost (USD)}} & \multicolumn{2}{c}{\shortstack{Estimated optimized\\retail cost (USD)}} & \multicolumn{2}{c}{Compactions} \\
\cmidrule(lr){2-3}\cmidrule(lr){4-5}\cmidrule(lr){6-7}\cmidrule(lr){8-9}\cmidrule(lr){10-11}
Configuration & Opus & GPT & Opus & GPT & Opus & GPT & Opus & GPT & Opus & GPT \\
\midrule
Baseline (no compaction) & 68.0 & 70.8 & 17.6\,/\,263.4\,/\,2.2 & 9.1\,/\,47.0\,/\,1.2 & \$274.70 & \$79.20 & \$274.70 & \$79.20 & 0\,/\,0\,/\,0 & 0\,/\,0\,/\,0 \\
Compaction, $16$K / $12$K & 67.6 & 72.6 & 20.4\,/\,61.4\,/\,2.1 & 9.6\,/\,49.5\,/\,1.6 & \$185.20 & \$90.20 & \$37.04 & \$18.04 & 1.4\,/\,1\,/\,9 & 0.1\,/\,0\,/\,3 \\
\bottomrule
\end{tabular}}}{
    \textbf{LongBench v2.}
    \emph{Accuracy} is the percentage answered correctly. \emph{Token spend} lists new input\,/\,cached input\,/\,output, expressed in M tokens, aggregated over the benchmark (\emph{new input} excludes
    prompt-cache hits).
    \emph{Compactions} are per task, given as mean\,/\,median\,/\,max.
    \emph{Nominal retail cost} uses the standard rates in \Cref{table:swe_profile}: per million new-input, cached-input, and output tokens, respectively, \$5, \$0.50, and \$25 for Opus~4.5; and \$4, \$0.40, and \$20 for GPT 5.6 Sol.
    \emph{Estimated optimized retail cost} applies the context factor in \Cref{appendix:retail_cost_fit} to the same token totals.
    \label{table:longbench_profile}
}
\end{table}

LOONG provides the clearest realistic multi-document result in support of our answer to \Cref{rq:long_input}: every setting in \Cref{table:loong_profile} improves both average score and perfect-answer rate for both models over the full-input baseline.
Depending on the setting, peak live context is bounded at $16$--$48$K tokens.
Compaction is correspondingly rare: both models have a median of zero compactions at the two larger settings, and even at $16$K / $12$K the medians are very low.

\begin{table}[H]
\centering
\captiontable{%
\resizebox{\textwidth}{!}{%
\begin{tabular}{l cc cc cc cc cc}
\toprule
 & \multicolumn{2}{c}{Accuracy (avg\,/\,perfect)} & \multicolumn{2}{c}{Token spend (M)} & \multicolumn{2}{c}{\shortstack{Nominal retail\\cost (USD)}} & \multicolumn{2}{c}{\shortstack{Estimated optimized\\retail cost (USD)}} & \multicolumn{2}{c}{Compactions} \\
\cmidrule(lr){2-3}\cmidrule(lr){4-5}\cmidrule(lr){6-7}\cmidrule(lr){8-9}\cmidrule(lr){10-11}
Configuration & Opus & GPT & Opus & GPT & Opus & GPT & Opus & GPT & Opus & GPT \\
\midrule
Baseline (no compaction) & 70.1\,/\,38.9 & 79.3\,/\,44.4 & 272.0\,/\,0.0\,/\,0.3 & 272.0\,/\,0.0\,/\,0.1 & \$1367.50 & \$1090.00 & \$1367.50 & \$1090.00 & 0\,/\,0\,/\,0 & 0\,/\,0\,/\,0 \\
Compaction, $48$K / $40$K & 79.8\,/\,52.6 & 84.6\,/\,53.0 & 57.6\,/\,232.1\,/\,5.2 & 23.7\,/\,85.1\,/\,6.2 & \$534.05 & \$252.84 & \$181.11 & \$75.85 & 0.5\,/\,0\,/\,11 & 0.2\,/\,0\,/\,10 \\
Compaction, $32$K / $24$K & 80.2\,/\,53.7 & 84.0\,/\,52.1 & 54.6\,/\,174.5\,/\,5.5 & 23.5\,/\,69.7\,/\,6.4 & \$497.75 & \$249.88 & \$134.18 & \$62.47 & 0.7\,/\,0\,/\,16 & 0.4\,/\,0\,/\,16 \\
Compaction, $16$K / $12$K & 79.3\,/\,53.4 & 82.7\,/\,51.9 & 55.4\,/\,115.2\,/\,6.7 & 31.1\,/\,61.0\,/\,13.7 & \$502.10 & \$422.80 & \$100.42 & \$84.56 & 1.4\,/\,1\,/\,46 & 2.0\,/\,0\,/\,61 \\
\bottomrule
\end{tabular}}}{\textbf{LOONG.} Columns and quantities are as in \Cref{table:longbench_profile}; \emph{Accuracy (avg\,/\,perfect)} lists average score\,/\,perfect-answer rate.\label{table:loong_profile}}
\end{table}

\Needspace{18\baselineskip}
AA-LCR shows the same benefit on multi-document reasoning (\Cref{table:aalcr_profile}).
At $16$K / $12$K, Claude Opus~4.5 reaches 84.0\% accuracy versus 65.0\% for the full-input baseline, while GPT 5.6 Sol reaches 83.0\% versus 82.0\%.

\begin{table}[H]
\centering
\captiontable{%
\resizebox{\textwidth}{!}{%
\begin{tabular}{l cc cc cc cc cc}
\toprule
 & \multicolumn{2}{c}{Accuracy} & \multicolumn{2}{c}{Token spend (M)} & \multicolumn{2}{c}{\shortstack{Nominal retail\\cost (USD)}} & \multicolumn{2}{c}{\shortstack{Estimated optimized\\retail cost (USD)}} & \multicolumn{2}{c}{Compactions} \\
\cmidrule(lr){2-3}\cmidrule(lr){4-5}\cmidrule(lr){6-7}\cmidrule(lr){8-9}\cmidrule(lr){10-11}
Configuration & Opus & GPT & Opus & GPT & Opus & GPT & Opus & GPT & Opus & GPT \\
\midrule
Baseline (no compaction) & 65.0 & 82.0 & 10.4\,/\,0.0\,/\,0.0 & 9.4\,/\,0.0\,/\,0.0 & \$52.00 & \$37.60 & \$52.00 & \$37.60 & 0.0\,/\,0\,/\,0 & 0.0\,/\,0\,/\,0 \\
Compaction, $32$K / $24$K & 80.0 & 83.0 & 2.7\,/\,17.2\,/\,0.3 & 0.9\,/\,3.0\,/\,0.3 & \$29.60 & \$10.80 & \$7.98 & \$2.70 & 0.2\,/\,0\,/\,3 & 0.1\,/\,0\,/\,3 \\
Compaction, $16$K / $12$K & 84.0 & 83.0 & 2.9\,/\,9.0\,/\,0.3 & 1.1\,/\,3.0\,/\,0.3 & \$26.50 & \$11.60 & \$5.30 & \$2.32 & 0.5\,/\,0\,/\,6 & 0.3\,/\,0\,/\,3 \\
\bottomrule
\end{tabular}}}{\textbf{AA-LCR.} Columns and quantities are as in \Cref{table:longbench_profile}.\label{table:aalcr_profile}}
\end{table}

On NIAH, selective input reading raises both models to 100\% accuracy (\Cref{table:niah_profile}).
The agents find the relevant evidence so directly that neither model performs a single compaction.
We attribute the stark success on this benchmark to existence of some more straightforward lexical overlap between the needles (and their known common synonyms) and the questions that are used to search for the needles.
Here we take the opportunity to also note that the entire concept of needle-in-a-haystack benchmarks is based on the premise that there is such a thing as a long-context task.
Starting from that premise, needle-in-a-haystack benchmarks are proposed as a proxy to evaluate model abilities to retrieve from such context as a part of operations the model under evaluation might have to perform in order to deal with the supposedly existing class of long-context tasks.
Again, we highlight that we have found no evidence of existence of inherently long-context tasks and have so far been able to address all tasks that have previously been producing long contexts by the simple means of selective input reading and compaction.

\begin{table}[H]
\centering
\captiontable{%
\resizebox{\textwidth}{!}{%
\begin{tabular}{l cc cc cc cc cc}
\toprule
 & \multicolumn{2}{c}{Accuracy} & \multicolumn{2}{c}{Token spend (M)} & \multicolumn{2}{c}{\shortstack{Nominal retail\\cost (USD)}} & \multicolumn{2}{c}{\shortstack{Estimated optimized\\retail cost (USD)}} & \multicolumn{2}{c}{Compactions} \\
\cmidrule(lr){2-3}\cmidrule(lr){4-5}\cmidrule(lr){6-7}\cmidrule(lr){8-9}\cmidrule(lr){10-11}
Configuration & Opus & GPT & Opus & GPT & Opus & GPT & Opus & GPT & Opus & GPT \\
\midrule
Baseline (no compaction) & 60.0 & 80.0 & 14.8\,/\,0.0\,/\,0.0 & 13.5\,/\,0.0\,/\,0.1 & \$74.00 & \$56.00 & \$74.00 & \$56.00 & 0\,/\,0\,/\,0 & 0\,/\,0\,/\,0 \\
Compaction, $48$K / $36$K & 100.0 & 100.0 & 0.8\,/\,0.0\,/\,0.0 & 0.2\,/\,0.0\,/\,0.0 & \$4.00 & \$0.80 & \$1.36 & \$0.24 & 0\,/\,0\,/\,0 & 0\,/\,0\,/\,0 \\
\bottomrule
\end{tabular}}}{\textbf{NIAH.} Columns and quantities are as in \Cref{table:longbench_profile}. The agent selectively reads from a filesystem, while the baseline processes the whole document.\label{table:niah_profile}}
\end{table}

\Needspace{20\baselineskip}
RULER further firmly supports our findings: mean recall rises from 62.5\% to 95.7\% for Claude Opus~4.5 and from 65.5\% to 96.4\% for GPT 5.6 Sol (\Cref{table:ruler_profile}).
Peak live context is bounded by a $16$K total context limit.
The median number of compactions is zero for both models, showing that selective input reading usually solves the tasks before any compactions are needed.

\begin{table}[H]
\centering
\captiontable{%
\resizebox{\textwidth}{!}{%
\begin{tabular}{l cc cc cc cc cc}
\toprule
 & \multicolumn{2}{c}{Accuracy (recall\,/\,exact)} & \multicolumn{2}{c}{Token spend (M)} & \multicolumn{2}{c}{\shortstack{Nominal retail\\cost (USD)}} & \multicolumn{2}{c}{\shortstack{Estimated optimized\\retail cost (USD)}} & \multicolumn{2}{c}{Compactions} \\
\cmidrule(lr){2-3}\cmidrule(lr){4-5}\cmidrule(lr){6-7}\cmidrule(lr){8-9}\cmidrule(lr){10-11}
Configuration & Opus & GPT & Opus & GPT & Opus & GPT & Opus & GPT & Opus & GPT \\
\midrule
Baseline (no compaction) & 62.5\,/\,51.3 & 65.5\,/\,54.9 & 175.5\,/\,0.0\,/\,0.1 & 176.9\,/\,0.0\,/\,0.9 & \$880.00 & \$725.60 & \$880.00 & \$725.60 & 0\,/\,0\,/\,0 & 0\,/\,0\,/\,0 \\
Compaction, $16$K / $12$K & 95.7\,/\,94.4 & 96.4\,/\,95.4 & 14.8\,/\,17.7\,/\,0.8 & 7.5\,/\,12.7\,/\,0.5 & \$102.85 & \$45.08 & \$20.57 & \$9.02 & 0.0\,/\,0\,/\,7 & 0.0\,/\,0\,/\,2 \\
\bottomrule
\end{tabular}}}{\textbf{RULER.} Columns and quantities are as in \Cref{table:longbench_profile}; \emph{Accuracy (recall\,/\,exact)} lists mean recall\,/\,exact-match rate.\label{table:ruler_profile}}
\end{table}

\Needspace{16\baselineskip}
On Counting-Stars, selective input reading reaches a 99.7\% mean score and 81.2\% exact match for both models, improving the exact-match rate from 65.1\% for Claude Opus~4.5 and 76.6\% for GPT 5.6 Sol (\Cref{table:counting_stars_profile}).
Neither model performs any compaction, so the result comes entirely from reading only the relevant input portions.

\begin{table}[H]
\centering
\captiontable{%
\resizebox{\textwidth}{!}{%
\begin{tabular}{l cc cc cc cc cc}
\toprule
 & \multicolumn{2}{c}{Accuracy (mean\,/\,exact)} & \multicolumn{2}{c}{Token spend (M)} & \multicolumn{2}{c}{\shortstack{Nominal retail\\cost (USD)}} & \multicolumn{2}{c}{\shortstack{Estimated optimized\\retail cost (USD)}} & \multicolumn{2}{c}{Compactions} \\
\cmidrule(lr){2-3}\cmidrule(lr){4-5}\cmidrule(lr){6-7}\cmidrule(lr){8-9}\cmidrule(lr){10-11}
Configuration & Opus & GPT & Opus & GPT & Opus & GPT & Opus & GPT & Opus & GPT \\
\midrule
Baseline (no compaction) & 98.5\,/\,65.1 & 99.3\,/\,76.6 & 9.5\,/\,0.0\,/\,0.3 & 9.5\,/\,0.0\,/\,0.3 & \$55.00 & \$44.00 & \$55.00 & \$44.00 & 0\,/\,0\,/\,0 & 0\,/\,0\,/\,0 \\
Compaction, $16$K / $12$K & 99.7\,/\,81.2 & 99.7\,/\,81.2 & 1.6\,/\,0.2\,/\,0.1 & 0.8\,/\,0.5\,/\,0.1 & \$10.60 & \$5.40 & \$2.12 & \$1.08 & 0\,/\,0\,/\,0 & 0\,/\,0\,/\,0 \\
\bottomrule
\end{tabular}}}{\textbf{Counting-Stars.} Columns and quantities are as in \Cref{table:longbench_profile}; \emph{Accuracy (mean\,/\,exact)} lists mean score\,/\,exact-match rate.\label{table:counting_stars_profile}}
\end{table}

BABILong combines selective input reading with a large accuracy gain (\Cref{table:babilong_profile}), where accuracy improves by 24.9 percentage points for Claude Opus~4.5 and 23.3 points for GPT 5.6 Sol.
Peak live context is bounded at $16$K tokens.
GPT 5.6 Sol never compacts, and Claude Opus~4.5 performs at most one compaction per task, once again showing that selective input reading does nearly all of the context reduction.

\begin{table}[H]
\centering
\captiontable{%
\resizebox{\textwidth}{!}{%
\begin{tabular}{l cc cc cc cc cc}
\toprule
 & \multicolumn{2}{c}{Accuracy} & \multicolumn{2}{c}{Token spend (M)} & \multicolumn{2}{c}{\shortstack{Nominal retail\\cost (USD)}} & \multicolumn{2}{c}{\shortstack{Estimated optimized\\retail cost (USD)}} & \multicolumn{2}{c}{Compactions} \\
\cmidrule(lr){2-3}\cmidrule(lr){4-5}\cmidrule(lr){6-7}\cmidrule(lr){8-9}\cmidrule(lr){10-11}
Configuration & Opus & GPT & Opus & GPT & Opus & GPT & Opus & GPT & Opus & GPT \\
\midrule
Baseline (no compaction) & 67.1 & 70.7 & 72.1\,/\,0.0\,/\,0.0 & 69.4\,/\,0.0\,/\,0.3 & \$360.50 & \$283.60 & \$360.50 & \$283.60 & 0\,/\,0\,/\,0 & 0\,/\,0\,/\,0 \\
Compaction, $16$K / $12$K & 92.0 & 94.0 & 8.3\,/\,6.4\,/\,0.5 & 2.3\,/\,2.3\,/\,0.3 & \$57.20 & \$16.12 & \$11.44 & \$3.22 & 0.0\,/\,0\,/\,1 & 0\,/\,0\,/\,0 \\
\bottomrule
\end{tabular}}}{\textbf{BABILong.} Columns and quantities are as in \Cref{table:longbench_profile}.\label{table:babilong_profile}}
\end{table}

\clearpage
\section{Agent outlines}
\label{appendix:agent_outlines}

The templates below give outlines of the structure of compaction prompts in \texttt{mini-swe-agent} and \texttt{mini-agent} used in this work.

\begingroup
\lstset{
  basicstyle=\ttfamily\footnotesize,
  breaklines=true,
  columns=fullflexible,
  keepspaces=true,
  showstringspaces=false,
  frame=single,
  postbreak={},
  aboveskip=6pt,
  belowskip=6pt
}

\subsection{\texttt{mini-swe-agent}}
\label{appendix:compaction_templates:swe}

We use the released \texttt{mini-swe-agent} harness \cite{minisweagent} v2.4, with negligible changes for environment integration and context management.
The underlying task-solving loop is fully retained.

To compact its earlier work, we use the following prompt passed as a user message once the compaction threshold is met.

\paragraph{Compaction prompt}
\begin{lstlisting}
Stop work. Write only a concise handoff recap; do not run commands or call tools.
The conversation will reset to the system prompt, task, and this recap.
{persistence}

Record:
1. Objective - one line describing the required result.
2. Established facts - exact paths, relevant symbols or commands, observed values,
   and important errors/output. Separate observations from unverified ideas.
3. Applied changes - each changed/created file, what changed, and whether verified.
4. Verification - exact checks run, outcomes, failures, obstacles, and workarounds.
5. Dead ends - failed approaches that should not be repeated.
6. Remaining plan - outstanding subgoals, in execution order; the immediate next
   command or edit; and the condition for completion.

{handoff_details}

If needed output was truncated, give a narrow command to retrieve it.
{submission_reminder}
\end{lstlisting}

\subsection{\texttt{mini-agent}}

As described in \Cref{section:methodology}, we generalize the base released \texttt{mini-swe-agent} harness into a \texttt{mini-agent} harness by removing SWE-specific instructions and adding instructions enabling additional classes of agentic workflows.

\paragraph{System prompt}
\begin{lstlisting}
You solve the supplied task by interacting with a computer through bash.
Work step by step: inspect relevant evidence, act, observe the result, and verify
that it meets the task's requirements. Keep command output focused.

{command_rules}

{context_instructions}
\end{lstlisting}

\paragraph{User prompt}
\begin{lstlisting}
<task>
{task}
</task>

{input_details}

Available tools:
{tools}

How to work:
{work_instructions}

Submission:
{submission_instructions}

{recap_section}
\end{lstlisting}

\paragraph{Compaction prompt}
\begin{lstlisting}
Stop work. Write only a concise handoff recap; do not call tools.
The conversation will reset, and the system prompt, task, and recap will be
provided again. {persistence}

Preserve what your next stretch needs:
1. Current progress and best answer/result, separating verified facts from assumptions.
2. Relevant evidence with exact values and source locations; changes already made,
   if any, with paths and verification status.
3. What was solved, what worked, what remains unresolved, what failed, and useful commands already tried.
4. Remaining plan - outstanding subgoals, in execution order; the immediate next
   command or edit; and the condition for completion.

{handoff_details}

If important output was truncated, name a narrow command to retrieve the missing
part. Do not invent observations or repeat the full task or document manifest.
{submission_reminder}
\end{lstlisting}

\noindent
\begin{tabularx}{\linewidth}{@{}lX@{}}
\toprule
\textbf{Variable} & \textbf{Description} \\
\midrule
\texttt{task} & Supplied question or task, including answer options and output requirements. \\
\texttt{input\_details} & Source filenames, document manifest, sizes, working directory, and other instance metadata. \\
\texttt{tools} & Available commands, arguments, and limits on retrieved text. \\
\texttt{work\_instructions} & Evidence-gathering procedure, calculations, verification requirements, and permitted changes. \\
\texttt{command\_rules} & Number of Bash calls required on work turns and shell execution behavior. \\
\texttt{submission\_instructions} & Exact completion command and required answer or artifact format. \\
\texttt{context\_instructions} & Instructions for resuming execution from the compacted state and producing the next compaction before starting a fresh conversation; empty without compaction. \\
\texttt{recap\_section} & Instructions for continuing execution, followed by the compacted state from the preceding conversation; empty without compaction. \\
\texttt{recap} & Compacted state inserted into \texttt{recap\_section} for the next conversation; initially a brief statement that no prior work exists. \\
\texttt{persistence} & Which files and environment state remain available across conversations, and which information must be preserved in the compacted state. \\
\texttt{handoff\_details} & Additional instructions specifying evidence, input coverage, calculations, or implementation state to preserve in the compaction. \\
\texttt{submission\_reminder} & Checks required before task completion and the completion protocol to preserve in the compacted state. \\
\bottomrule
\end{tabularx}

\endgroup

\end{document}